%% file: egpaper_for_review.tex
\documentclass[10pt,twocolumn,letterpaper]{article}

\usepackage{cvpr}
\usepackage{times}
\usepackage{epsfig}
\usepackage{graphicx}
\usepackage{amsmath}
\usepackage{amssymb}

\usepackage[pagebackref=true,breaklinks=true,letterpaper=true,colorlinks,bookmarks=false]{hyperref}

\usepackage{gensymb,graphics,subfigure,inputenc}
\usepackage[linesnumbered,algo2e,boxed]{algorithm2e}
\graphicspath{{Figure/}}
\usepackage[figuresright]{rotating}
\usepackage{amsmath,amssymb,textcomp,subfigure,multirow,upgreek,bm}
\usepackage{booktabs}
\usepackage{color,mdwlist}
\usepackage{graphicx}

\usepackage{xcolor}
\usepackage{enumitem}
\setitemize{noitemsep,topsep=0pt,parsep=0pt,partopsep=0pt}

\usepackage{caption}
\usepackage{flushend}

\usepackage{enumitem}
\setitemize{noitemsep,topsep=0pt,parsep=0pt,partopsep=0pt}

\def\etal{et al.~}
 
\cvprfinalcopy 

\def\cvprPaperID{5211} 

\begin{document}

\title{Local Class-Specific and Global Image-Level Generative Adversarial Networks for Semantic-Guided Scene Generation}

\author{Hao Tang$^1$ \quad Dan Xu$^2$ \quad Yan Yan$^3$ \quad Philip H. S. Torr$^2$ \quad Nicu Sebe$^{1,4}$ \\
$^1$University of Trento \quad $^2$University of Oxford \quad $^3$Texas State University \quad $^4$Huawei Research Ireland
}

\maketitle

\input{0Abstract}

\input{1Introduction}
\input{2RelatedWork}

\input{3Formulation}

\input{4Implementation}
\input{5Conclusion}

\clearpage
{\small
	\bibliographystyle{ieee_fullname}
	\bibliography{egbib}
}

\clearpage
\input{supplementary}

\end{document}

%% file: 0Abstract.tex
\begin{abstract}

In this paper, we address the task of semantic-guided scene generation. One open challenge widely observed in global image-level generation methods is the difficulty of generating small objects and detailed local texture. To tackle this issue, in this work we consider learning the scene generation in a local context, and correspondingly design a local class-specific generative network with semantic maps as a guidance, which separately constructs and learns sub-generators concentrating on the generation of different classes, and is able to provide more scene details. To learn more discriminative class-specific feature representations for the local generation, a novel classification module is also proposed. To combine the advantage of both global image-level and the local class-specific generation, a joint generation network is designed with an attention fusion module and a dual-discriminator structure embedded.
Extensive experiments on two scene image generation tasks show superior generation performance of the proposed model. State-of-the-art results are established by large margins on both tasks and on challenging public benchmarks. The source code and trained models are available at
\url{https://github.com/Ha0Tang/LGGAN}.

\end{abstract}

%% file: 1Introduction.tex
\section{Introduction}

\begin{figure}[!t] \small
	\centering
	\includegraphics[width=1\linewidth]{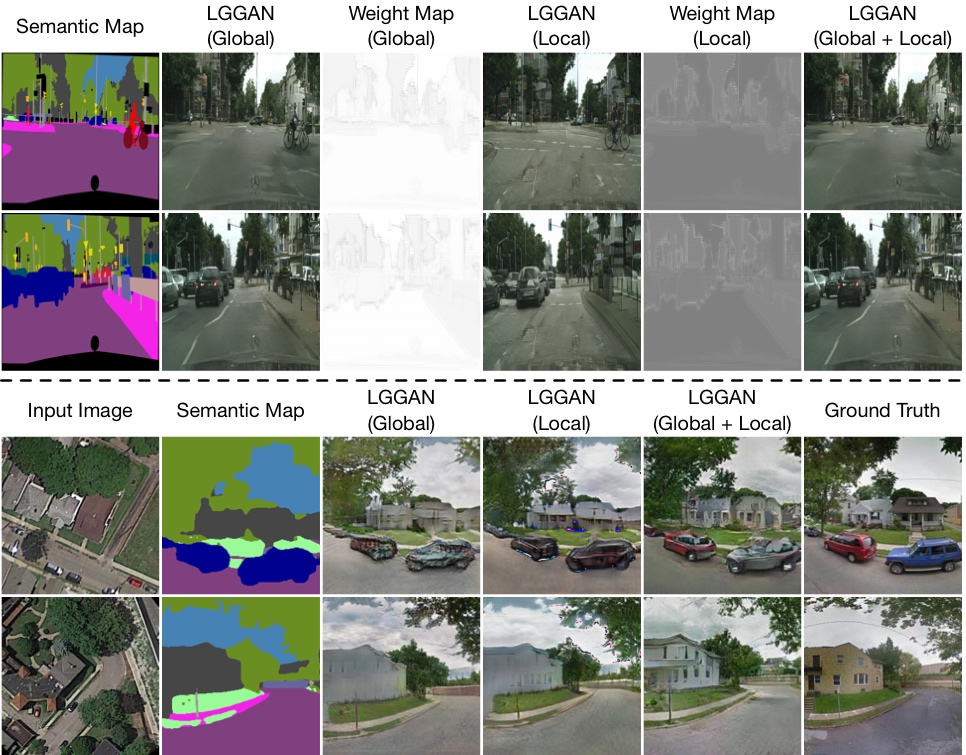}
	\caption{Examples of semantic image synthesis results on Cityscapes  (\textit{top}) and cross-view image translation results on Dayton (\textit{bottom}) with different settings of our LGGAN. 
		}
	\label{fig:first}
	\vspace{-0.4cm}
\end{figure}

\begin{figure*}[!t] \small
	\centering
	\includegraphics[width=1\linewidth]{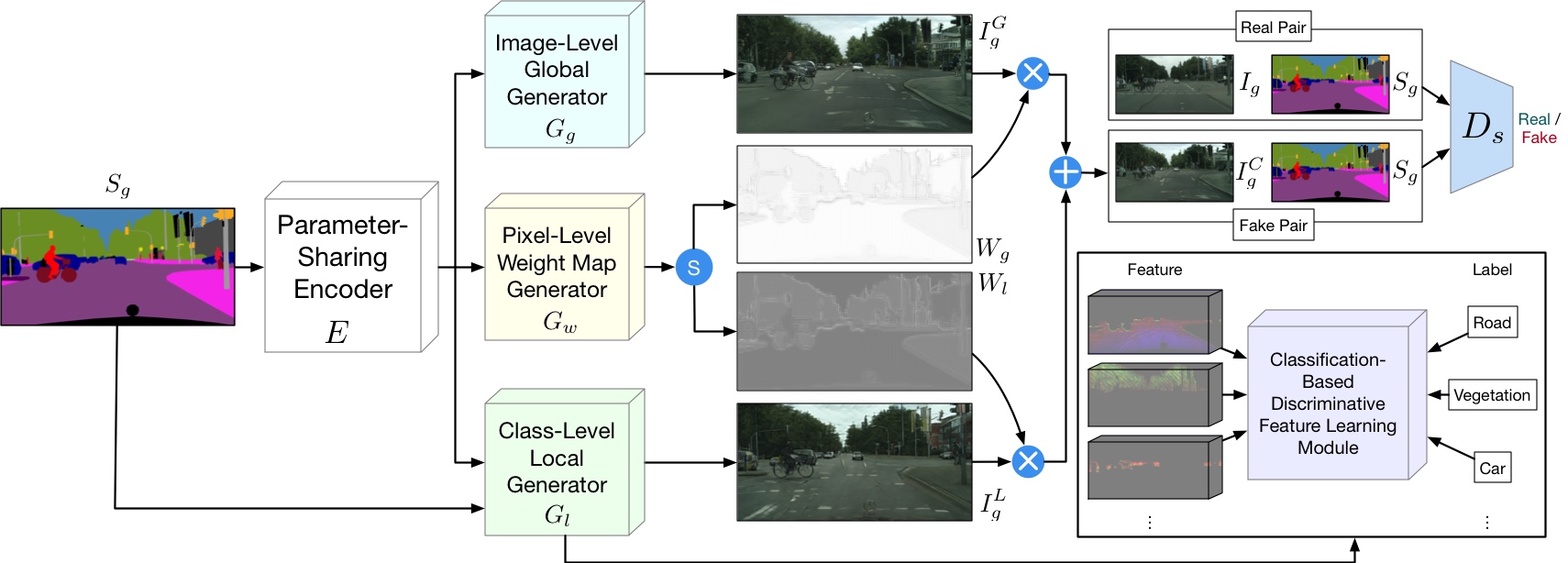}
	\caption{Overview of the proposed LGGAN, which contains a semantic-guided generator $G$ and  discriminator $D_s$. $G$ consists of a parameter-sharing encoder $E$, an image-level global generator $G_g$, a class-level local generator $G_l$ and a weight map generator $G_w$.
	The global generator and local generator are automatically combined by two learned weight maps from the weight map generator to reconstruct the target image.
	$D_s$ tries to distinguish the generated images from two modality spaces, i.e., image space and semantic space. 
	Moreover, to learn a more discriminative class-specific feature representation, a novel classification module is proposed. 
	All of these components are trained in an end-to-end fashion so that the local generation and the global generation can benefit from each other.
	The symbols $\oplus$, $\otimes$ and $\textcircled{s}$ denote element-wise addition, element-wise multiplication and channel-wise Softmax, respectively.}
	\label{fig:framework}
	\vspace{-0.5cm}
\end{figure*}

Semantic-guided scene generation is a hot research topic covering several main-stream research directions, including cross-view image translation \cite{isola2017image,zhai2017predicting,regmi2018cross,regmi2019cross,tang2019multi,regmi2019bridging} and semantic image synthesis~\cite{wang2018high, chen2017photographic, qi2018semi, park2019semantic}. 
The cross-view image translation task proposed in \cite{regmi2018cross} is essentially an ill-posed problem due to the large ambiguity in the generation if only a single RGB image is given as input. 
To alleviate this problem, recent works such as SelectionGAN~\cite{tang2019multi} try to generate the target image based on an image of the scene and several novel semantic maps, as shown in Fig.~\ref{fig:first}(\textit{bottom}).
Adding a semantic map allows the model to learn the correspondences in the target view with appropriate object relations and transformations.
On the other side, the semantic image synthesis task aims to generate a photo-realistic image from a semantic map 
\cite{wang2018high, chen2017photographic, qi2018semi, park2019semantic}, as shown in Fig.~\ref{fig:first}(\textit{top}). Recently, Park et al.~\cite{park2019semantic} propose a spatially-adaptive normalization for synthesizing photo-realistic images given an input semantic map.
With the useful semantic information, existing methods on both tasks achieved promising performance in scene generation.

However, one can still observe unsatisfying perspectives, especially on the generation of local scene structure and details as well as small scale objects, which we believe are mainly due to several reasons.
First, existing methods on both tasks are mostly based on a global image-level generation, which accepts a semantic map containing several object classes and aims to generate the appearance of all the different classes, by using the same network design or using shared network parameters. In this case, all the classes are treated equally by the network. While different semantic classes have distinct properties, specific network learning for different semantic classes intuitively would benefit the complex multi-class generation. 
Second, we observe that the number of training samples of different scene classes is imbalanced.
For instance, for the Dayton dataset~\cite{vo2016localizing}, the cars and  buses only occupy less than 2\% with respect to all pixels in the training data, which naturally makes the model learning be dominated by the classes with the larger number of training samples. 
Third, the size of objects in different scene classes is diverse. As shown in the first row of Fig.~\ref{fig:first}, 
larger-scale object classes such as road, sky usually occupy bigger area of the image than smaller-scale classes such as pole and traffic light. Since the convolutional network usually shares the parameters at different convolutional positions, the larger-scale object classes would thus take advantage during the learning, further increasing the difficult in generating well the small-scale object classes.

\par To tackle these issues, a straightforward consideration would be to model the generation of different scene classes specifically in a local context. By so doing, each class could have its own generation network structure or parameters, thus greatly avoiding the learning of a biased generation space. To achieve this goal, in this paper we design a novel class-specific generation network. It consists of several sub-generators for different scene classes with a shared encoded feature map. The input semantic map is utilized as the guidance to obtain feature maps corresponding to each class spatially, which are then used to produce a separate generation for different class regions.  

\par Due to the highly complementary properties of global and local generation, a Local class-specific and Global image-level Generative Adversarial Network~(LGGAN) is proposed to combine the advantage of these two. It mainly contains three network branches (see Fig.~\ref{fig:framework}). The first branch is the image-level global generator, which learns a global appearance distribution using the input, and the second branch is the proposed class-specific local generator, which aims to generate different objects classes separately using semantic-guided class-specific feature filtering. Finally, the fusion weight-map generation branch learns two pixel-level weight maps which are used to fuse the local and global sub-networks in a weighted-combination of their final generation results. The proposed LGGAN can be jointly trained in an end-to-end fashion to make the local and global generation benefit each other in the optimization. 


Overall, the contributions of this paper are as follows:
\begin{itemize}[leftmargin=*]
	\item We explore scene generation from the local context, which we believe is beneficial to generate richer scene details compared with the existing global image-level generation methods. A new local class-specific generative structure has been designed for this purpose. It can effectively handle the generation of small objects and scene details which are common difficulties encountered by the global-based generation.
	\item We propose a novel global and local generative adversarial network design able to take into account both the global and local contexts. To stabilize the optimization of the proposed joint network structure, a fusion weight-map generator and a dual-discriminator are introduced. Moreover, to learn discriminative class-specific feature representations, a novel classification module is proposed.
	\item Experiments for cross-view image translation on the Dayton~\cite{vo2016localizing} and CVUSA~\cite{workman2015wide} datasets,
	and semantic image synthesis on the Cityscapes~\cite{cordts2016cityscapes} and ADE20K~\cite{zhou2017scene} datasets demonstrate the effectiveness of the proposed LGGAN framework, and show significantly better results compared with state-of-the-art methods on both tasks. 
\end{itemize}

%% file: 2RelatedWork.tex
\section{Related Work}
\par\noindent\textbf{Generative Adversarial Networks (GANs)}
\cite{goodfellow2014generative} have been widely used for image generation \cite{karras2017progressive,zhang2018self,brock2018large,karras2018style,gulrajani2017improved,goetschalckx2019ganalyze,shocher2019ingan,liu2019few,shaham2019singan}.
A vanilla GAN has two important components, i.e., a generator and a discriminator. The goal of the generator is to generate photo-realistic images from a noise vector, while the discriminator is trying to distinguish between the real and the generated image. 
To synthesize user-specific images, Conditional GAN (CGAN) \cite{mirza2014conditional} has been proposed.
A CGAN combines a vanilla GAN and an external information, such as class labels~\cite{odena2016semi,odena2016conditional,choi2017stargan}, text descriptions \cite{li2019manigan,han2017stackgan,li2019controllable}, object keypoint~\cite{reed2016learning,tang2019cycle}, human body/hand skeleton~\cite{albahar2019guided,tang2018gesturegan,balakrishnan2018synthesizing,zhu2019progressive}, conditional images~\cite{zhu2017unpaired,isola2017image}, semantic maps~\cite{wang2018high,tang2019multi,park2019semantic,wang2018video}, scene graphs~\cite{johnson2018image,zhao2019image,ashual2019specifying} and attention maps~\cite{zhang2018self,mejjati2018unsupervised,tang2019attentiongan}.

\par\noindent\textbf{Global and Local Generation in GANs.}
Modelling global and local information in GANs to generate better results has been used in various generative tasks \cite{huang2017beyond,iizuka2017globally,lin2019coco,li2018global,qi2018global,gu2019mask}. 
For instance, Huang et al.~\cite{huang2017beyond} propose TPGAN for frontal view synthesis by simultaneously perceiving global structures and local details.
Gu et al.~\cite{gu2019mask} propose MaskGAN for face editing by separately learning every face component, e.g., mouth and eye.
However, these methods are only applied to face-related tasks such as face rotation or face editing, where all the domains have large  overlap and similarity.
However, we propose a new local and global image generation framework design for a more challenging scene generation task, and the local context modeling is based on semantic-guided class-specific generation, which is not explored by any existing works.

\noindent\textbf{Scene Generation.}
Scene generation tasks are a hot topic as each image can be parsed into distinctive semantic objects~\cite{bau2019seeing,ashual2019specifying,turkoglu2019layer,gong2019autogan,bau2019semantic,bau2018gan}.
In this paper, we mainly focus on two scene generation tasks, i.e., cross-view image translation \cite{zhai2017predicting,regmi2018cross,regmi2019cross,tang2019multi} and semantic image synthesis \cite{wang2018high, chen2017photographic, qi2018semi, park2019semantic}.
Most existing works on cross-view image translation have been conducted to synthesize novel views of the same objects  \cite{dosovitskiy2017learning,zhou2016view,tatarchenko2016multi,choy20163d}.
Moreover, several works deal with image translation problems with drastically different views and generate a novel scene from a given different scene ~\cite{zhai2017predicting,regmi2018cross,regmi2019cross,tang2019multi}.
For instance,
Tang~\etal \cite{tang2019multi} propose SelectionGAN to solve the cross-view image translation task using semantic maps and CGAN models.
On the other side, the semantic image synthesis task aims to generate a photo-realistic image from a semantic map~\cite{wang2018high, chen2017photographic, qi2018semi, park2019semantic}.
For example, Park et al. propose GauGAN~\cite{park2019semantic}, which achieves the best results on this task.

With the semantic maps as guidance, existing approaches on both tasks achieve promising performance.
However, we still observe that the results produced by these global image-level generation methods are often unsatisfactory, especially on detailed local texture.
In contrast, our proposed approach focuses on generating a more realistic global structure/layout and local texture details.
Both local and global generation branches are jointly learned in an end-to-end fashion that aims at using the mutually improved benefits from each other.


%% file: 3Formulation.tex
\section{The Proposed LGGAN}
We start by presenting the details of the proposed Local class-specific and Global image-level GANs (LGGAN). An illustration of the overall framework is shown in Fig.~\ref{fig:framework}. 
The generation module mainly consists of three parts, i.e., a semantic-guided class-specific generator modelling the local context, an image-level generator modelling the global layout, and a weight-map generator for fusing the local and the global generators. 
We first introduce the used backbone structure, and then present the design of the proposed local and global generation networks.

\subsection{The Backbone Encoding Network Structure}
\noindent \textbf{Semantic-Guided Generation.}
In this paper, we mainly focus on two tasks, i.e., semantic image synthesis and cross-view image translation.
For the former, we follow GauGAN~\cite{park2019semantic} and use the semantic map $S_g$ as the input of the backbone encoder $E$, as shown in Fig.~\ref{fig:framework}.
For the latter, we follow SelectionGAN~\cite{tang2019multi} and concatenate the input image $I_a$ and a novel semantic map $S_g$ as the input of the backbone encoder $E$.
By so doing, the semantic maps act as priors to guide the model to learn the generation of another domain.

\noindent \textbf{Parameter-Sharing Encoder.}
As we have three different branches for three different generators, the encoder $E$ is sharing parameters to all the three branches to make a compact backbone network. 
The gradients from all the three branches contribute together to the learning of the encoder. 
We believe that in this way, the encoder can learn both local and global information and the correspondence between them.
Then the encoded deep representations from the input $S_g$ can be represented as $E(S_g)$, as shown in Fig.~\ref{fig:framework}.

\begin{figure*}[!t]
	\centering
	\includegraphics[width=0.88\linewidth]{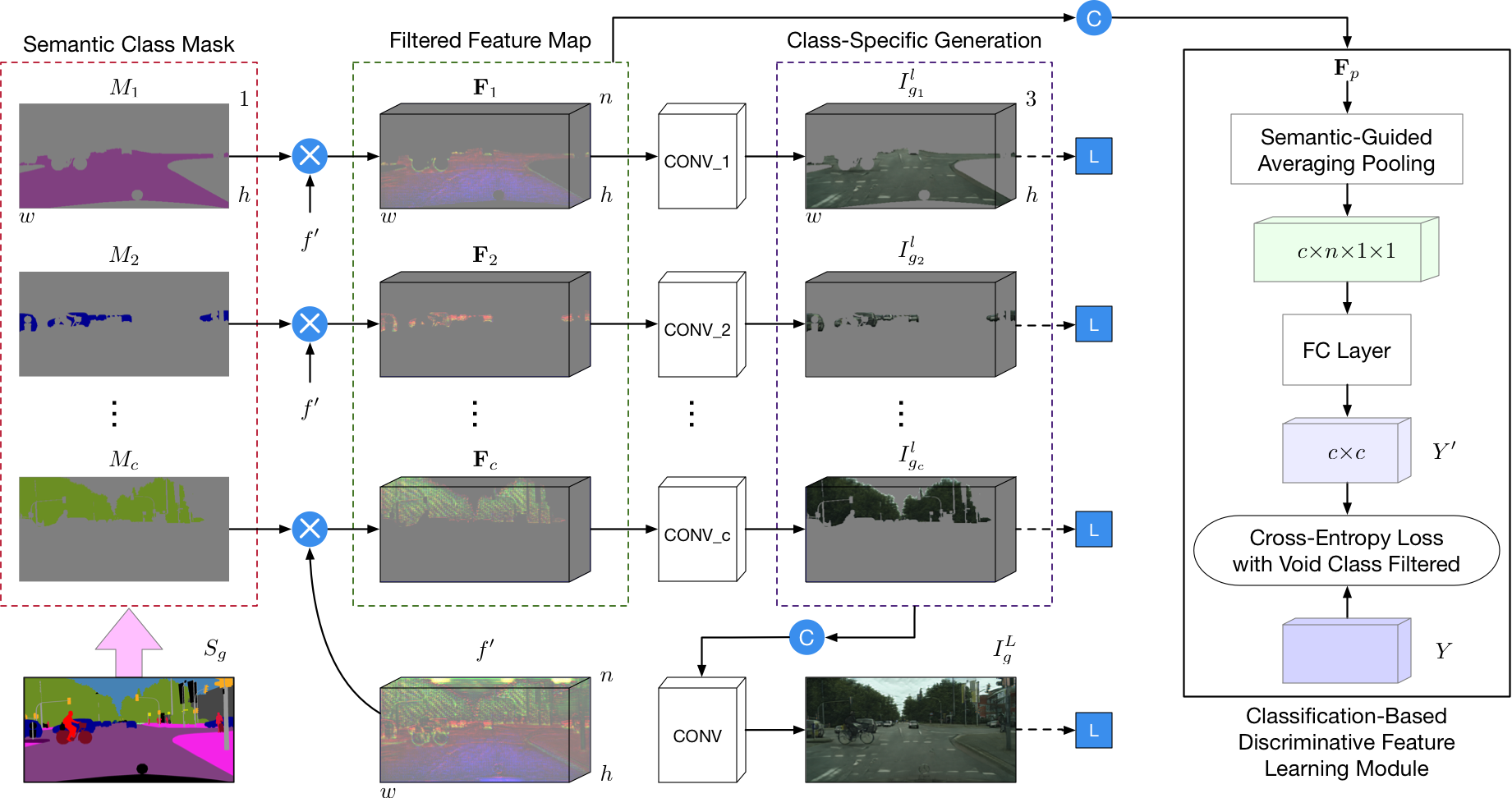}
	\caption{Overview of the proposed local class-specific generator $G_l$ consisting of four steps, i.e., semantic class mask calculation, class-specific feature map filtering, classification-based discriminative feature learning and class-specific generation. A cross-entropy loss with void class filtered is applied at each class feature representation for learning a more discriminative class-specific feature representation. A semantic-mask guided pixel-wise $L1$ loss is applied at the end for class-level reconstruction. 	The symbols $\otimes$ and $\textcircled{c}$ denote  element-wise multiplication and channel-wise concatenation, respectively.
	}
	\label{fig:local}
	\vspace{-0.4cm}
\end{figure*}

\subsection{The LGGAN Structure}
\par\noindent\textbf{Class-Specific Local Generation Network.}
As shown in Fig.~\ref{fig:first} and discussed in the introduction, the issue of training data imbalance between different classes and size difference between scene objects makes it extremely difficult to generate small object classes and scene details. 
To overcome this limitation, we propose a novel local class-specific generation network design. It separately constructs a generator for each semantic class being thus able to largely avoid the interference from the large object classes in the joint optimization. 
Each sub-generation branch has independent network parameters and concentrates on a specific class, being therefore capable of effectively producing similar generation quality for different classes and yielding richer local scene details. 
\par The overview of the local generation network $G_l$ is illustrated in Fig.~\ref{fig:local}. The encoded features $E(S_g)$ are first fed into two consecutive deconvolutional layers to increase the spatial size  with the number of channels reduced two times. Then the scaled feature map $f'$ is multiplied by the semantic mask of each class, i.e., $M_i$, to obtain a filtered class-specific feature map for each one. The mask-guided feature filtering operation can be written as: 
\begin{equation}
\mathrm{\textbf{F}}_i = M_i * f',  \quad i=1,2,...,c,
\label{eqn:fea}
\end{equation}
where $c$ is the number of semantic classes. 
Then the filtered feature map $\mathrm{\textbf{F}}_i$ is fed into several convolutional layers for the corresponding $i^{th}$ class and generate an output image $I_{g_i}^l$. For better learning each class, we utilize a semantic-mask guided pixel-wise $L1$ reconstruction loss, which can be expressed as follows:
\begin{equation}
\begin{aligned}
\mathcal{L}_{\mathrm{L1}}^{local} = \sum_{i=1}^{c} \mathbb{E}_{I_g, I_{g_i}^l} \lbrack \vert\vert I_g* M_i - I_{g_i}^l \vert\vert_1 \rbrack.
\label{eqn:classl1}
\end{aligned}
\end{equation}
The final output $I_g^L$ from the local generation network can be obtained in two ways. 
The first one is performing an element-wise addition of all the class-specific outputs:
\begin{equation}
I_g^L = I_{g_1}^l \oplus I_{g_2}^l \oplus \cdots \oplus I_{g_c}^l.
\label{eq:local}
\end{equation}
The second one is performing a convolutional operation on all the class-specific outputs, as shown in Fig.~\ref{fig:local},
\begin{equation}
I_g^L = \mathrm{Conv}(\mathrm{Concat}(I_{g_1}^l, I_{g_2}^l, \cdots, I_{g_c}^l)),
\label{eq:local2}
\end{equation}
where $\mathrm{Concat}(\cdot)$ and $\mathrm{Conv}(\cdot)$ denote channel-wise concatenation and convolutional operation, respectively.


\noindent \textbf{Class-Specific Discriminative Feature Learning.}
We observe that the filtered feature map $\mathbf{F}_i$ is not able to produce very discriminative class-specific generations, leading to similar generation results for some classes, especially for small-scale object classes.
In order to have more diverse generation for different object classes, we propose a novel classification-based feature learning module to learn more  discriminative class-specific feature representations, as shown in Fig.~\ref{fig:local}.
One input sample of the module is a pack of feature maps produced from different local generation branches, i.e.,~$\{\mathbf{F}_1, ..., \mathbf{F}_c\}$.
First, the packed feature map $\mathbf{F}_p {\in} \mathbb{R}^{c {\times} n {\times} h {\times} w}$ (with $n, h, w$ as the number of feature map channels, height and width, respectively) is fed into a semantic-guided averaging pooling layer, and we obtain a pooled feature map with dimension of $c {\times} n {\times} 1 {\times} 1$.
Then the pooled feature map is connected with a fully connected layer to predict classification probability of the $c$ object classes of the scene. 
The output after the FC layer is $Y^{'} {\in} \mathbb{R}^{c {\times} c}$, where $c$ is the number of semantic classes, as for each filtered feature map $\mathbf{F}_i$ ($i{=}1,...,c$), we predict a $c {\times} 1$ one-hot vector for the probabilities of the $c$ classes.

Since some object classes may not exist in the input semantic mask sample, the features from the local branches corresponding to the void classes should not contribute to the classification loss. Therefore, we filter the final cross-entropy loss by multiplying it with a void class indicator for each input sample. The indicator is an one hot vector $H {=} \{H_i\}_{i=1}^{c}$ with $H_i {=} 1$ for a valid class and $H_i{=}0$ for a void class. Then, the Cross-Entropy (CE) loss is defined as follows: 
\begin{equation}
\mathcal{L}_{\mathrm{CE}} = -\sum_{m=1}^{c} H_m\sum_{i=1}^{c}1\{Y(i)=i\} \log(f(\mathbf{F}_i)),
\label{eq:class_loss}
\end{equation}
where $1\{\cdot\}$ is an indicator function, i.e., having a return 1 if $Y(i){=}i$ else 0.  $f(\cdot)$ is a classification function which produces a prediction probability given an input feature map $\mathbf{F}_i$. $Y$ is a label set of all the object classes.

\par\noindent\textbf{Image-Level Global Generation Network.} Similar to the local generation branch, $E(S_g)$ is also fed into the global generation sub-network $G_g$ for global image-level generation, as shown in Fig.~\ref{fig:framework}. Global generation is capable to capture the global structure information or layout of the targeted images. 
Thus, the global result $I_g^G$ can be obtained through a feed-forward computation: $I_g^G {=} G_g (E(S_g)).$
Besides the proposed $G_g$, many existing global generators can also be used together with the proposed local generator $G_l$, making the proposed framework very flexible.


\begin{table*}[!h] \small
	\centering
	\caption{Quantitative evaluation of the Dayton dataset in the a2g direction. For all metrics except KL score, higher is better. 
($\ast$) Inception Score for real (ground truth) data is 3.8319, 2.5753 and 3.9222 for all, top-1 and top-5 setups, respectively. 
}
	\resizebox{0.83\linewidth}{!}{%
		\begin{tabular}{lcccccccccccc} \toprule
			\multirow{2}{*}{Method} & \multicolumn{4}{c}{Accuracy (\%)} & \multicolumn{3}{c}{Inception Score$^\ast$} & \multirow{2}{*}{SSIM} & \multirow{2}{*}{PSNR} & \multirow{2}{*}{SD} & \multirow{2}{*}{KL}  \\ \cmidrule(lr){2-5} \cmidrule(lr){6-8} 
			& \multicolumn{2}{c}{Top-1} & \multicolumn{2}{c}{Top-5} & All & Top-1 & Top-5 \\ \hline
			Pix2pix \cite{isola2017image}          &6.80 &9.15 &23.55&27.00& 2.8515&1.9342&2.9083 & 0.4180 &17.6291&19.2821& 38.26 $\pm$ 1.88 \\
			X-SO \cite{regmi2019cross} & 27.56 & 41.15 & 57.96 & 73.20 & 2.9459 & 2.0963 & 2.9980 & 0.4772 & 19.6203 & 19.2939 & 7.20 $\pm$ 1.37 \\
			X-Fork \cite{regmi2018cross}           &30.00&48.68&61.57&78.84& 3.0720&2.2402&3.0932 &0.4963&19.8928&19.4533  &6.00 $\pm$ 1.28 \\
			X-Seq \cite{regmi2018cross}               & 30.16&49.85&62.59&80.70& 2.7384&2.1304&2.7674 &0.5031 &20.2803 &19.5258 & 5.93 $\pm$ 1.32 \\ \cmidrule(lr){1-12}
			Pix2pix++~\cite{isola2017image} &32.06 &54.70&63.19&81.01&3.1709&2.1200&3.2001&0.4871&21.6675&18.8504& 5.49 $\pm$ 1.25\\
			X-Fork++~\cite{regmi2018cross} &34.67 & 59.14 &66.37&84.70&3.0737&2.1508&3.0893&0.4982&21.7260&18.9402& 4.59 $\pm$ 1.16 \\
			X-Seq++~\cite{regmi2018cross} & 31.58 & 51.67 &65.21 & 82.48 &3.1703&2.2185&3.2444&0.4912&21.7659&18.9265& 4.94 $\pm$ 1.18 \\
			SelectionGAN~\cite{tang2019multi} & 42.11 & 68.12 & 77.74 & 92.89 & 3.0613 & 2.2707 & 3.1336 & \textbf{0.5938} & \textbf{23.8874} & \textbf{20.0174} & 2.74 $\pm$ 0.86 \\
			LGGAN (Ours) & \textbf{48.17} & \textbf{79.35} & \textbf{81.14} & \textbf{94.91} & \textbf{3.3994} & \textbf{2.3478} & \textbf{3.4261} & 0.5457& 22.9949 & 19.6145 & \textbf{2.18 $\pm$ 0.74}\\
			\bottomrule		
	\end{tabular}}
	\label{tab:dayton}
	\vspace{-0.4cm}
\end{table*}

\begin{table*}[!tbp] \small
	\centering
	\caption{Quantitative evaluation of the CVUSA dataset in a2g direction. For all metrics except KL score, higher is better. ($\ast$) Inception Score for real (ground truth) data is 4.8741, 3.2959 and 4.9943 for all, top-1 and top-5 setups, respectively.}
	\resizebox{0.83\linewidth}{!}{%
		\begin{tabular}{lccccccccccccc} \toprule
			\multirow{2}{*}{Method}  & \multicolumn{4}{c}{Accuracy (\%)}& \multicolumn{3}{c}{Inception Score$^\ast$} & \multirow{2}{*}{SSIM} & \multirow{2}{*}{PSNR} & \multirow{2}{*}{SD} & \multirow{2}{*}{KL} \\ \cmidrule(lr){2-5} \cmidrule(lr){6-8} 
			& \multicolumn{2}{c}{Top-1} & \multicolumn{2}{c}{Top-5} & All & Top-1 & Top-5  \\ \hline
			Zhai \etal \cite{zhai2017predicting}   &13.97 &14.03 &42.09 &52.29 & 1.8434 &1.5171  &1.8666  & 0.4147 &17.4886 &16.6184   & 27.43 $\pm$ 1.63  \\
			Pix2pix \cite{isola2017image} &7.33  &9.25  &25.81 &32.67  & 3.2771 &2.2219 &3.4312  & 0.3923  &17.6578  &18.5239  & 59.81 $\pm$ 2.12   \\ 
			X-SO \cite{regmi2019cross} & 0.29 & 0.21  & 6.14  & 9.08  & 1.7575  & 1.4145  & 1.7791  & 0.3451  & 17.6201  & 16.9919  & 414.25 $\pm$ 2.37 \\
			X-Fork \cite{regmi2018cross}           &20.58 &31.24 &50.51 &63.66  &3.4432 &2.5447 &3.5567  & 0.4356  &19.0509  &18.6706  & 11.71 $\pm$ 1.55 \\
			X-Seq \cite{regmi2018cross}             &15.98 &24.14 &42.91 &54.41  &3.8151 &2.6738 &\textbf{4.0077}  & 0.4231  &18.8067  &18.4378  &15.52 $\pm$ 1.73  \\ \hline
			Pix2pix++ \cite{isola2017image} & 26.45 & 41.87 & 57.26 & 72.87 & 3.2592 & 2.4175 & 3.5078 & 0.4617 & 21.5739 & 18.9044 & 9.47 $\pm$ 1.69 \\
			X-Fork++ \cite{regmi2018cross} &31.03 &49.65 &64.47 &81.16 &3.3758&2.5375&3.5711&0.4769 & 21.6504& 18.9856 &7.18 $\pm$ 1.56\\
			X-Seq++ \cite{regmi2018cross} &34.69&54.61&67.12&83.46&3.3919&2.5474&3.4858 & 0.4740 & 21.6733 & 18.9907 & 5.19 $\pm$ 1.31\\
			SelectionGAN~\cite{tang2019multi} & 41.52  & 65.51  & 74.32  & 89.66 & 3.8074  & 2.7181 & 3.9197  & \textbf{0.5323}  & \textbf{23.1466}  & 19.6100  & 2.96 $\pm$ 0.97  \\ 
			LGGAN (Ours) & \textbf{44.75} & \textbf{70.68} & \textbf{78.76} & \textbf{93.40} & \textbf{3.9180} & \textbf{2.8383} & 3.9878 & 0.5238 & 22.5766 & \textbf{19.7440} & \textbf{2.55 $\pm$ 0.95}  \\            	
			\bottomrule		
		\end{tabular}}
		\label{tab:cvusa}
					\vspace{-0.4cm}
	\end{table*}

\par\noindent\textbf{Pixel-Level Fusion Weight-Map Generation Network.} In order to better combine the local and the global generation sub-networks, we further propose a pixel-level weight map generator $G_w$, which aims at predicting pixel-wise weights for fusing the global generation $I_g^G$ and the local generation $I_g^L$. 
In our implementation, $G_w$ consists of two Transpose Convolution$\rightarrow$InstanceNorm$\rightarrow$ReLU blocks and one Convolution$\rightarrow$InstanceNorm$\rightarrow$ReLU block.
The number of the output channels for these three block are 128, 64 and~2, respectively. The kernel sizes are $3{\times}3$ with stride~2 except for the last layer which has a kernel size of $1{\times}1$ with stride~1 for dense prediction. 
We predict a two-channel weight map $W_f$ using the following calculation:
\begin{equation}
W_f = \mathrm{Softmax}(G_w(E(S_g))),
\label{eqn:final}
\end{equation}
where $\mathrm{Softmax}(\cdot)$ denotes a channel-wise softmax function used for normalization, i.e., the sum of the weight values at the same pixel position is equal to 1. By so doing, we can guarantee that information from the combination would not explode. $W_f$ is sliced to have a weight map $W_g$ for the global branch and a weight map $W_l$ for the local branch. The fused final generation result is calculated as follows:
\begin{equation}
I_g^C = I_g^G \otimes W_g + I_g^L \otimes W_l,
\label{eqn:final_fusion}
\end{equation}
where $\otimes$ is an element-wise multiplication operation.
In this way, the pixel-level weights predicted from $G_w$ directly operate on the output of $G_g$ and $G_l$.
Moreover, generators $G_w$, $G_g$ and $G_l$ affect and contribute to each other in the model optimization. 

\par\noindent\textbf{Dual-Discriminator.} To exploit the prior domain knowledge, i.e., the semantic map, we extend the single domain vanilla discriminator~\cite{goodfellow2014generative} to a cross domain structure and we refer to it as the semantic-guided discriminator $D_s$, as shown in Fig.~\ref{fig:framework}. It employs the input semantic map $S_g$ and the generated image $I_g^C$ (or the real image $I_g$) as input: 
\begin{equation}
\begin{aligned}
\mathcal{L}_{\mathrm{CGAN}}(G, D_s) = 
& \mathbb{E}_{S_g, I_g} \left[ \log D_s(S_g, I_g) \right] +  \\
& \mathbb{E}_{S_g, I_g^C} \left[\log (1 - D_s(S_g, I_g^C)) \right],
\end{aligned}
\label{eqn:conditonalgan2}
\end{equation}
which aims to preserve scene layout and capture the local-aware information.


For the cross-view image translation task, we also propose another image-guided discriminator $D_i$, which takes the conditional image $I_a$ and the final generated image $I_g^C$ (or the ground-truth image $I_g$) as input:
\begin{equation}
\begin{aligned}
\mathcal{L}_{\mathrm{CGAN}}(G, D_i) = 
& \mathbb{E}_{I_a, I_g} \left[ \log D_i(I_a, I_g) \right] +  \\
& \mathbb{E}_{I_a, I_g^C} \left[\log (1 - D_i(I_a, I_g^C)) \right].
\end{aligned}
\label{eqn:conditonalgan1}
\end{equation}
In this case, the total loss of our Dual-Discriminator $D$ is $\mathcal{L}_{\mathrm{CGAN}} {=} \mathcal{L}_{\mathrm{CGAN}}(G, D_i) {+} \mathcal{L}_{\mathrm{CGAN}}(G, D_s)$.

%% file: 4Implementation.tex
\section{Experiments}
\label{sec:experiments}
The proposed LGGAN can be applied to different generative tasks such as the cross-view image translation~\cite{tang2019multi} and the semantic image synthesis~\cite{park2019semantic}. In this section, we present experimental results and analysis on both tasks.

\subsection{Results on Cross-View Image Translation}

\noindent \textbf{Datasets and Evaluation Metric.}
We follow \cite{tang2019multi} and perform cross-view image translation experiments on both Dayton~\cite{vo2016localizing} and CVUSA datasets~\cite{workman2015wide}.
Similarly to \cite{regmi2018cross,tang2019multi}, we employ Inception Score (IS), Accuracy (Acc.), KL Divergence Score (KL), Structural-Similarity (SSIM), Peak Signal-to-Noise Ratio (PSNR) and Sharpness Difference (SD) to evaluate the proposed model.

\noindent\textbf{State-of-the-Art Comparisons.}
We compare our LGGAN with several recently proposed state-of-the-art methods, i.e., 
Zhai et al.~\cite{zhai2017predicting},
Pix2pix~\cite{isola2017image}, X-SO~\cite{regmi2019cross}, X-Fork~\cite{regmi2018cross} and X-Seq~\cite{regmi2018cross}.
The comparison results are shown in Tables~\ref{tab:dayton} and ~\ref{tab:cvusa}.
We can observe that LGGAN consistently outperforms the competing methods on all metrics. 

\begin{figure}[!t] \small
	\centering
	\includegraphics[width=1\linewidth]{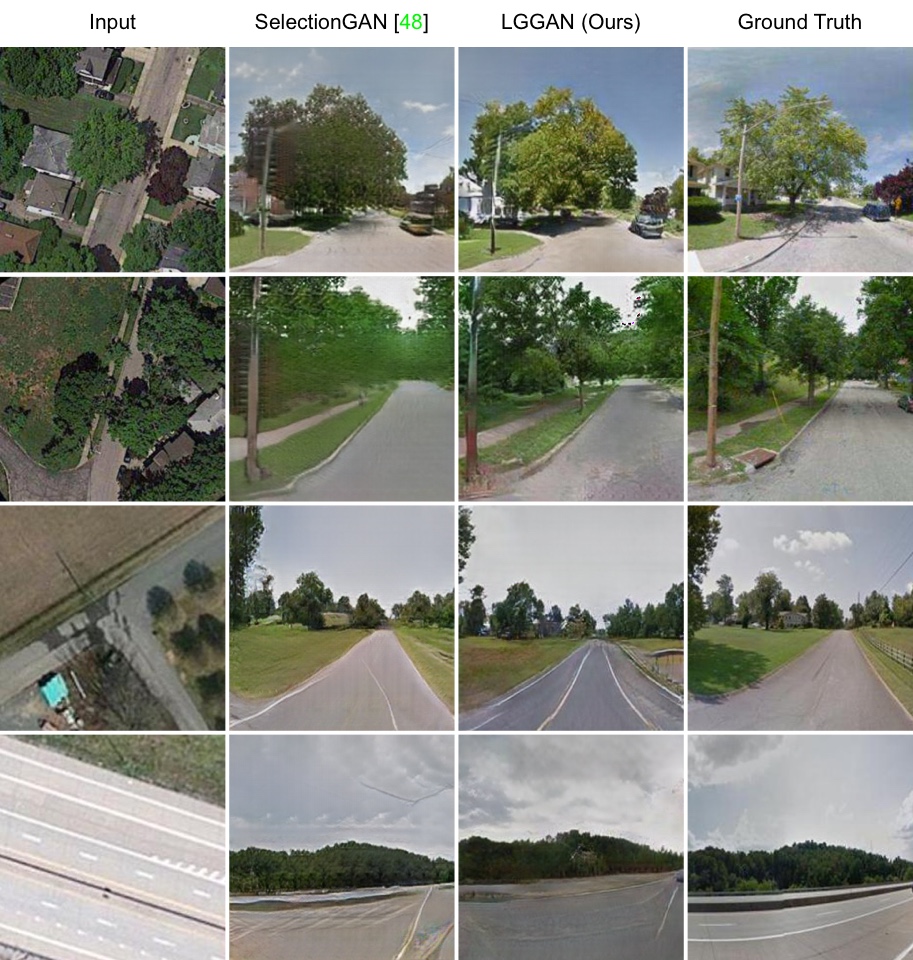}
	\caption{Qualitative comparison in a2g direction on Dayton (\textit{top two rows}) and CVUSA (\textit{bottom two rows}).
	}
	\label{fig:dayton256}
	\vspace{-0.4cm}
\end{figure}

To study the effectiveness of LGGAN, we conduct experiments with the methods using semantic maps and RGB images as input, including Pix2pix++~\cite{isola2017image}, X-Fork++~\cite{regmi2018cross}, X-Seq++~\cite{regmi2018cross} and SelectionGAN~\cite{tang2019multi}.
We implement Pix2pix++, X-Fork++ and X-Seq++ using their public source code.
Results are shown in Tables~\ref{tab:dayton} and~\ref{tab:cvusa}. 
We observe that LGGAN achieves significantly better results than Pix2pix++, X-Fork++ and X-Seq++, confirming the advantage of the proposed LGGAN.
A direct comparison with SelectionGAN is also shown in the tables providing better results on most metrics except pixel-level evaluation metrics, i.e., SSIM, PSNR and SD. 
SelectionGAN uses a two-stage generation strategy and an attention selection module, achieving slightly better results than ours on these three metrics.
However, we generate much more photo-realistic results than SelectionGAN as shown in Fig.~\ref{fig:dayton256}.

\noindent\textbf{Qualitative Evaluation.}
The qualitative results compared with the leading method SelectionGAN \cite{tang2019multi} are shown in Fig.~\ref{fig:dayton256}.
We observe that the  results generated by the proposed LGGAN are visually better than SelectionGAN.
Specifically, our method generates more clear details on objects such as cars, buildings, road, trees than SelectionGAN.

\begin{figure*}[!t] \small
	\centering
	\includegraphics[width=1\linewidth]{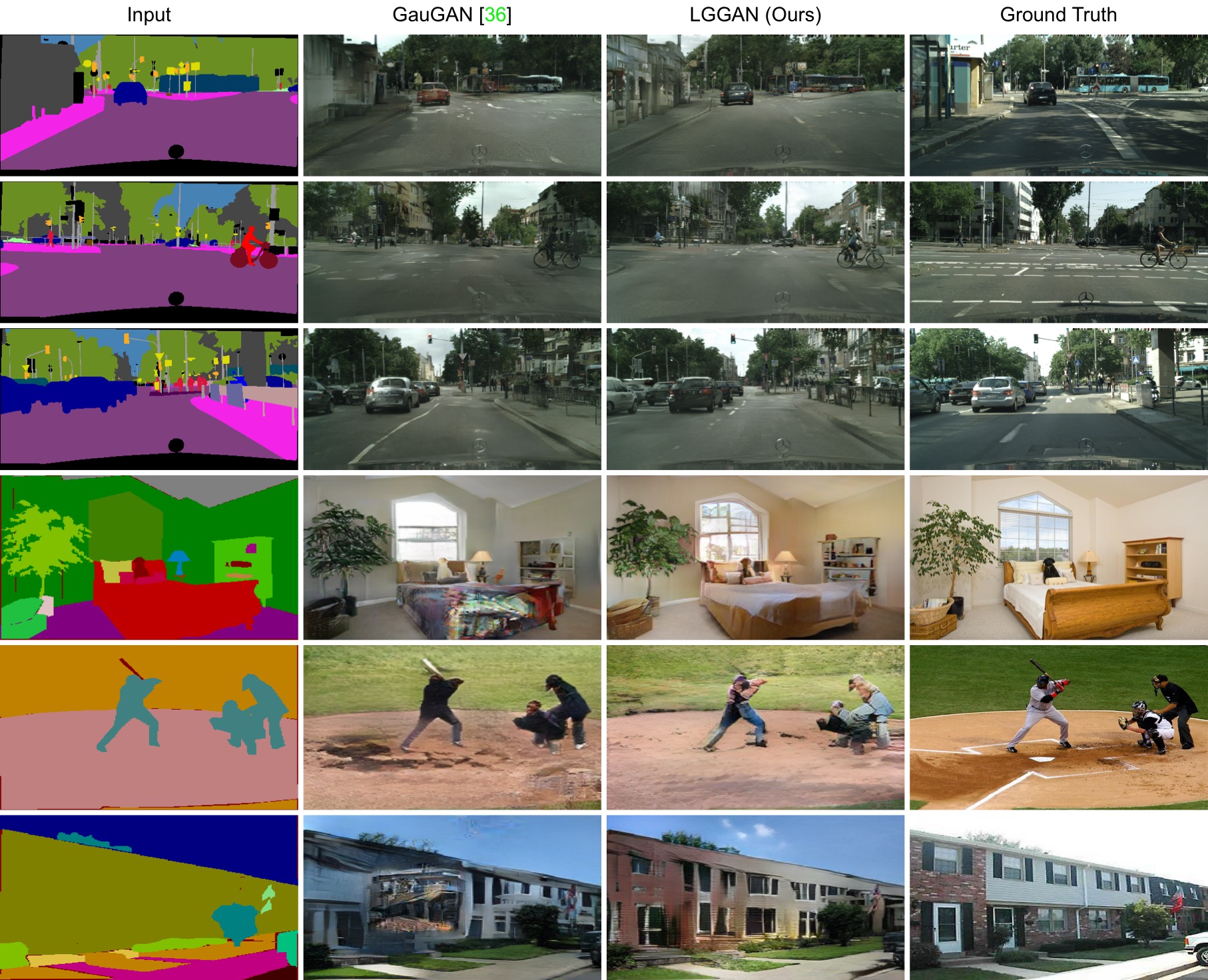}
	\caption{Qualitative comparison  on Cityscapes (\textit{top three rows}) and ADE20K (\textit{bottom three rows}).
	}
	\label{fig:cityscapes}
	\vspace{-0.3cm}
\end{figure*}

\begin{table*}[!t]\small
	\centering
	\caption{(\textit{left}) Our method significantly outperforms current leading methods in semantic segmentation scores (mIoU and Acc) and FID. (\textit{middle}) User preference study. The numbers indicate the percentage of users who favor the results of the proposed LGGAN over the competing method. (\textit{right}) Quantitative comparison of different variants of the proposed LGGAN on the semantic image synthesis tasks. For mIoU, Acc and AMT, higher is better. For FID, lower is better. 
	}
	\resizebox{0.41\linewidth}{!}{%
		\begin{tabular}{lcccccc} \toprule
			\multirow{2}{*}{Method}  & \multicolumn{3}{c}{Cityscapes} & \multicolumn{3}{c}{ADE20K} \\ \cmidrule(lr){2-4} \cmidrule(lr){5-7} 
			      & mIoU $\uparrow$    & Acc $\uparrow$  & FID  $\downarrow$ & mIoU $\uparrow$    & Acc $\uparrow$  & FID  $\downarrow$  \\ \midrule
			CRN~\cite{chen2017photographic}  & 52.4  & 77.1 & 104.7  & 22.4 & 68.8 & 73.3 \\
			SIMS~\cite{qi2018semi}           & 47.2  & 75.5 & \textbf{49.7} & N/A & N/A & N/A\\
			Pix2pixHD~\cite{wang2018high}    & 58.3  & 81.4 & 95.0  & 20.3 & 69.2 & 81.8 \\
			GauGAN~\cite{park2019semantic}   & 62.3  & 81.9 & 71.8  & 38.5 & 79.9 & 33.9 \\
			LGGAN (Ours)      & \textbf{68.4} & \textbf{83.0} & 57.7 & \textbf{41.6} & \textbf{81.8} & \textbf{31.6} \\	\bottomrule
		\end{tabular}}
		\hspace{0.2cm}
    	\resizebox{0.29\linewidth}{!}{%
		\begin{tabular}{lcc} \toprule
        AMT $\uparrow$ & Cityscapes                      & ADE20K  \\ \midrule
        Ours vs. CRN~\cite{chen2017photographic} & 67.38 & 79.54 \\
        Ours vs. Pix2pixHD~\cite{wang2018high} & 56.16   & 85.69 \\ 
        Ours vs. SIMS~\cite{qi2018semi} & 54.84          & N/A   \\
        Ours vs. GauGAN~\cite{park2019semantic} & 53.19  & 57.31 \\ \bottomrule
        \end{tabular}} 
        \hspace{0.2cm}
        \resizebox{0.25\linewidth}{!}{%
        \begin{tabular}{lcc} \toprule
			Setup of LGGAN &  mIoU $\uparrow$ & FID $\downarrow$  \\ \midrule	
			S1: Ours w/ Global                                         &  62.3   & 71.8  \\
			S2: S1 + Local (add.)         &  64.6    & 66.1   \\	
			S3: S1 + Local (con.)    &  65.8     & 65.6  \\ 
			S4: S3 + Class Dis. Loss &  67.0 & 61.3  \\
			S5: S4 + Weight Map &  \textbf{68.4} & \textbf{57.7} \\ \bottomrule
		\end{tabular}}
	\label{tab:results}
	\vspace{-0.4cm}
\end{table*}

\subsection{Results on Semantic Image Synthesis}
\noindent \textbf{Datasets and Evaluation Metric}. 
We follow \cite{park2019semantic} and conduct extensive experiments on both Cityscapes~\cite{cordts2016cityscapes} and ADE20K~\cite{zhou2017scene} datasets.
We use the mean Intersection-over-Union (mIoU), pixel accuracy (Acc) and Fr\'echet Inception Distance (FID)~\cite{heusel2017gans} as the evaluation metrics.

\noindent \textbf{State-of-the-Art Comparisons.}
We compare the proposed LGGAN with several leading semantic image synthesis methods, i.e., Pix2pixHD~\cite{wang2018high}, CRN~\cite{chen2017photographic}, SIMS~\cite{qi2018semi} and GauGAN~\cite{park2019semantic}.
Results of the mIoU, Acc and FID metrics are shown in Table~\ref{tab:results}(\textit{left}).
We find that the proposed LGGAN outperforms the existing competing methods by a large margin on both mIoU and Acc metrics.
For FID, the proposed method is only worse than SIMS on Cityscapes.
However, SIMS has poor segmentation performance.
The reason is that SIMS produces an image by searching and copying image patches from the training dataset. 
The generated images are more realistic since the method uses the real image patches.
However, the approach always tends to copy objects with mismatched patches due to queries that cannot be guaranteed to have results in the dataset.
Moreover, we follow the evaluation protocol of GauGAN and provide AMT results, as shown in Table~\ref{tab:results}(\textit{middle}).
We observe that users favor our synthesized results on both datasets compared with other competing methods including SIMS.

\begin{figure*}[!t] \small
	\centering
	\includegraphics[width=1\linewidth]{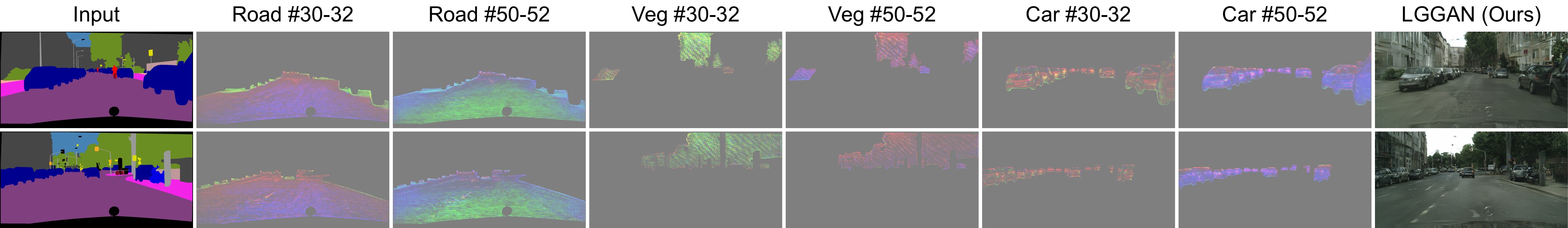}
	\caption{Visualization of learned class-specific feature maps on 3 different classes, i.e., road, vegetation and car. 
	}
	\label{fig:feature}
	\vspace{-0.4cm}
\end{figure*}

\begin{figure*}[!t] \small
	\centering
	\includegraphics[width=1\linewidth]{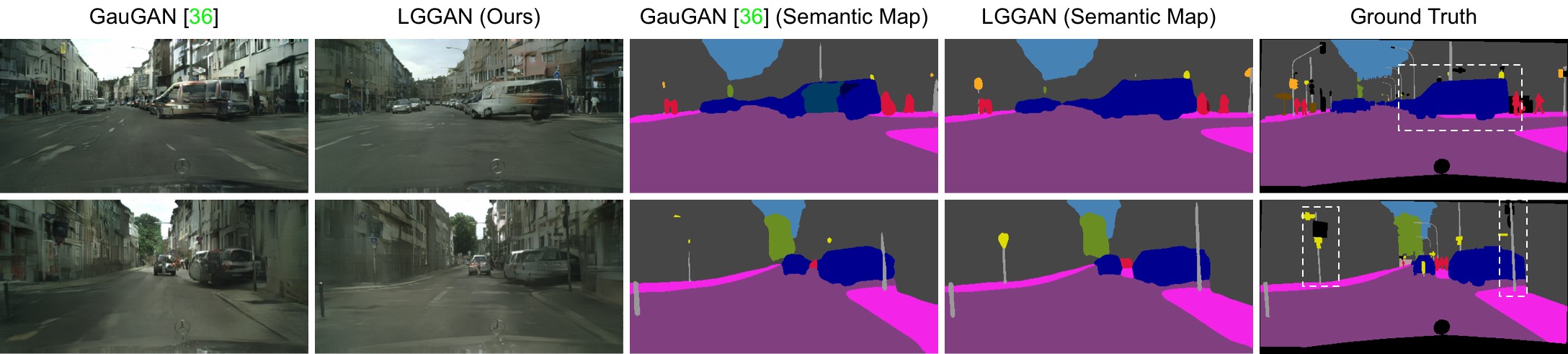}
	\caption{Visualization of generated semantic maps compared with those from GauGAN~\cite{park2019semantic} on Cityscapes. 
	}
	\label{fig:seg}
	\vspace{-0.4cm}
\end{figure*}

\noindent \textbf{Qualitative Evaluation.}
The qualitative results compared with the leading method GauGAN \cite{park2019semantic} are shown in Fig.~\ref{fig:cityscapes}.
We can see that the proposed LGGAN generates much better results with fewer visual artifacts than GauGAN.

\par\noindent\textbf{Visualization of Learned Feature Maps.}
In Fig.~\ref{fig:feature}, we randomly show some channels from the learned class-specific feature maps (30$^{th}$ to 32$^{th}$, and 50$^{th}$ to 52$^{th}$) on Cityscapes to see if they clearly highlight particular semantic classes.
We show the visualization results on 3 different classes, i.e.,~road, vegetation and car.
We can easily observe that each local sub-generator learns well the class-level deep representations, further confirming our motivations.

\par\noindent \textbf{Visualization of Generated Semantic Maps.}
We follow GauGAN~\cite{park2019semantic} and apply pretrained segmentation networks on the generated images to produce semantic maps, i.e., DRN-D-105~\cite{yu2017dilated} for Cityscapes and UperNet101~\cite{xiao2018unified} for ADE20K.
The generated semantic maps of the proposed LGGAN, GauGAN and the ground truths are shown in Fig.~\ref{fig:seg}.
We observe that our LGGAN generates better semantic maps than GauGAN, especially on local texture (`car' in the first row) and small objects (`traffic sign' and `pole' in the second row), confirming our initial motivation.

\subsection{Ablation Study}
We conduct extensive ablation studies on the Cityscapes dataset to evaluate different components of our LGGAN.

\noindent \textbf{Baseline Models.}
The proposed LGGAN has 5 baselines (i.e., S1, S2, S3, S4, S5) as shown in Table~\ref{tab:results}(\textit{right}): 
(i) S1 means only adopting the global generator.
(ii) S2 combines the global generator and the proposed local generator to produce the final results, in which the local results are produced by using an addition operation as proposed in Eq.~\eqref{eq:local}.
(iii) The difference between S3 and S2 is that S3 uses 
a convolutional layer to generate the local results, as presented in Eq.~\eqref{eq:local2}.
(iv) S4 employ the proposed classification-based discriminative feature learning module.
(v) S5 is our full model and adopts the proposed weight map fusion strategy.

\noindent \textbf{Effect of Local and Global Generation.}
The results of the ablation study are shown in Table~\ref{tab:results}(\textit{right}).
When using an addition operation to generate the local result, the local and global generation strategy improves mIoU and FID by 2.3 and 5.7, respectively.
When adopting a convolutional operation to produce the local results, the performance boosts further, i.e., 3.5 and 6.2 gain on the mIoU and FID metrics, respectively.
Both results confirm the effectiveness of the proposed local and global generation framework.
We also provide qualitative results of the local and global generation in Fig.~\ref{fig:first}. 
We observe that our full model, i.e., Global + Local,  generates visually better results than both the global and local method, 
which further confirms our motivations.

\noindent \textbf{Effect of Classification-Based Feature Learning.}
S4 significantly outperforms S3 with around 1.2 and 4.3 gain on the mIoU and FID metric, respectively. This means that the model indeed learns a more discriminative class-specific feature representation, confirming our design motivation.

\noindent \textbf{Effect of Weight Map Fusion.}
By adding the proposed weight map fusion scheme in S5, the overall performance is further boosted with 1.4 and 3.6 improvement on the mIoU and FID metric, respectively.
This means the proposed LGGAN indeed learns complementary information from the local and the global generation branch.
In Fig.~\ref{fig:first}, we show some samples of the generated weight maps. 

%% file: 5Conclusion.tex
\section{Conclusion}
We propose Local class-specific and Global image-level Generative Adversarial Networks (LGGAN) for semantic-guided scene generation. 
The proposed LGGAN contains three generation branches, i.e., global image-level generation, local class-level generation and pixel-level fusion weight map generation, respectively.
A new class-specific local generation network is designed to alleviate the influence of imbalanced training data and size difference of scene objects in joint learning. 
To learn more class-specific discriminative feature representations, a novel classification module is further proposed. 
Experimental results demonstrate the superiority of the proposed approach and show new state-of-the-art results on both cross-view image translation and semantic image synthesis tasks.

\noindent \textbf{Acknowledgments.} This work was supported by the Italy-China collaboration project
TALENT: CN19GR09.

%% file: supplementary.tex
This document  provides additional implementation details and experimental results on the semantic image synthesis task. 
First, we provide detailed
training details about the proposed LGGAN (Sec.~\ref{sec:2}). 
Second, we compare the proposed LGGAN with  state-of-the-art methods (Sec.~\ref{sec:4}), i.e., Pix2pixHD~\cite{wang2018high}, CRN~\cite{chen2017photographic}, SIMS~\cite{qi2018semi} and GauGAN~\cite{park2019semantic}.
Additionally, we compare our `Local + Global'
with `Global' and `Local' model (Sec.~\ref{sec:3}). 
We also provide the visualization
results of the generated semantic maps (Sec.~\ref{sec:6}). 
Finally, we show some failure cases of the proposed  LGGAN on this task (Sec.~\ref{sec:7}).

\section{More Results on Semantic Image Synthesis}


\subsection{Training Details}
\label{sec:2}
Following GauGAN~\cite{park2019semantic}, all the experiments are conducted on an NVIDIA DGX-1 with 8 V100 GPUs.
We perform 200 epochs of training on both  Cityscapes~\cite{cordts2016cityscapes} and ADE20K~\cite{zhou2017scene} datasets.
The images are resized to $512 {\times} 256$ and $256 {\times} 256$ on Cityscapes and ADE20K, respectively.
We adopt the Spectral Norm~\cite{miyato2018spectral} to all the layers in both the generator and discriminator.
We also apply the Adam~\cite{kingma2014adam} optimizer with momentum terms  $\beta_1{=}0$ and $\beta_2{=}0.999$ as our solver.

\subsection{State-of-the-art Comparison}
\label{sec:4}
\begin{figure*}[!ht] 
	\centering
	\includegraphics[width=1\linewidth]{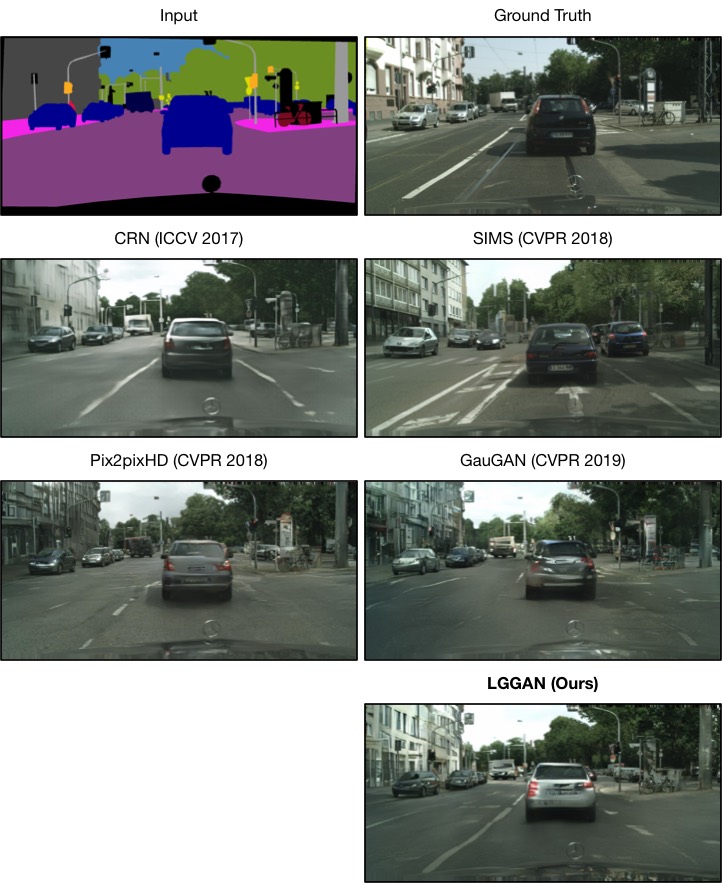}
	\caption{Results of the comparison with those from the Pix2pixHD~\cite{wang2018high}, CRN~\cite{chen2017photographic}, SIMS~\cite{qi2018semi} and GauGAN~\cite{park2019semantic} methods on the Cityscapes dataset. These samples were randomly selected without cherry-picking for visualization purposes.
	}
	\label{fig:city1}
\end{figure*}

\begin{figure*}[!ht] 
	\centering
	\includegraphics[width=1\linewidth]{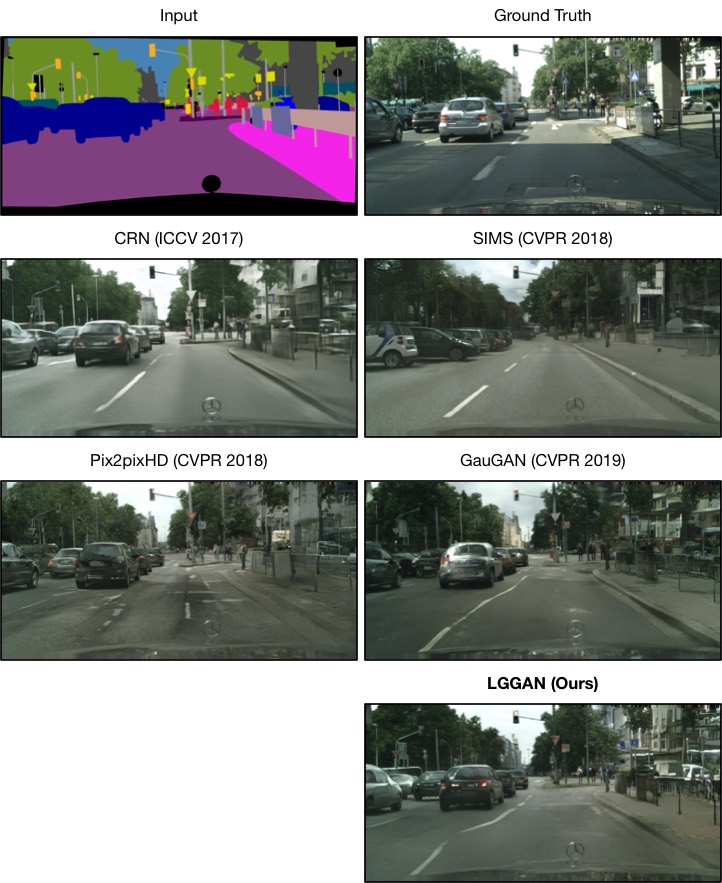}
	\caption{Results of the comparison with those from the Pix2pixHD~\cite{wang2018high}, CRN~\cite{chen2017photographic}, SIMS~\cite{qi2018semi} and GauGAN~\cite{park2019semantic} methods on the Cityscapes dataset. These samples were randomly selected without cherry-picking for visualization purposes.
	}
	\label{fig:city2}
\end{figure*}

\begin{figure*}[!ht] 
	\centering
	\includegraphics[width=1\linewidth]{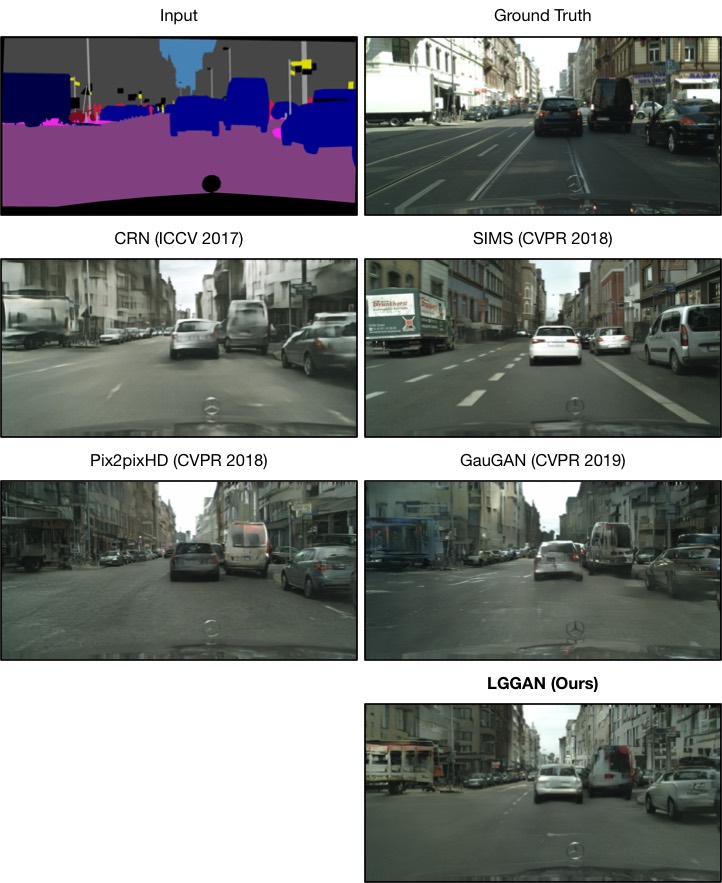}
	\caption{Results of the comparison with those from the Pix2pixHD~\cite{wang2018high}, CRN~\cite{chen2017photographic}, SIMS~\cite{qi2018semi} and GauGAN~\cite{park2019semantic} methods on the Cityscapes dataset. These samples were randomly selected without cherry-picking for visualization purposes.
	}
	\label{fig:city3}
\end{figure*}

\begin{figure*}[!ht] 
	\centering
	\includegraphics[width=1\linewidth]{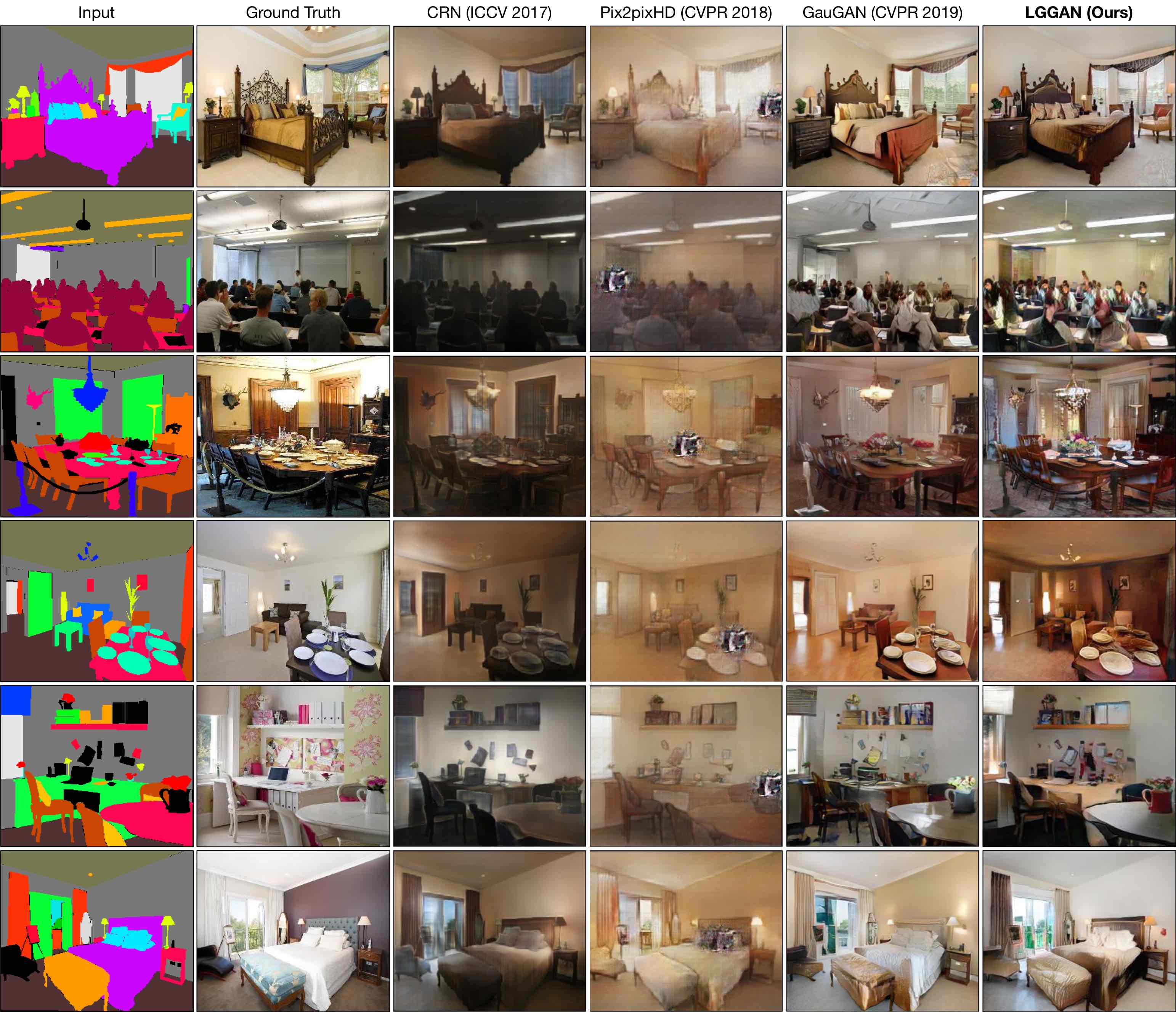}
	\caption{Results of the comparison with those from the Pix2pixHD~\cite{wang2018high}, CRN~\cite{chen2017photographic} and GauGAN~\cite{park2019semantic} methods on the ADE20K dataset. These samples were randomly selected without cherry-picking for visualization purposes.
	}
	\label{fig:ade20k_1}
\end{figure*}

\begin{figure*}[!ht] 
	\centering
	\includegraphics[width=1\linewidth]{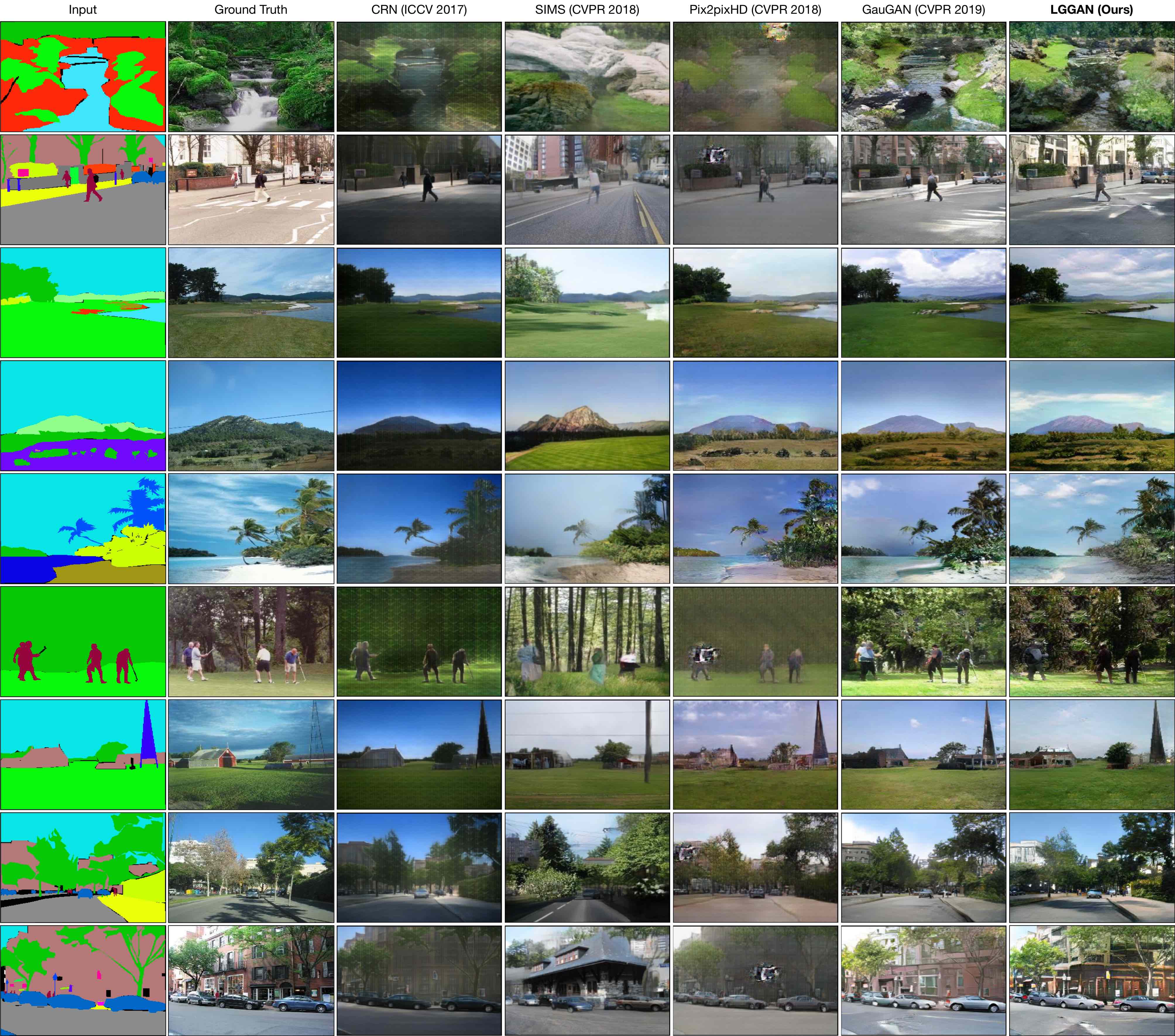}
	\caption{Results of the comparison with those from the Pix2pixHD~\cite{wang2018high}, CRN~\cite{chen2017photographic}, SIMS~\cite{qi2018semi} and GauGAN~\cite{park2019semantic} methods on the ADE20K dataset. These samples were randomly selected without cherry-picking for visualization purposes.
	}
	\label{fig:ade20k_2}
\end{figure*}

We show more generation results of the proposed LGGAN on both datasets compared with those from the leading models, i.e., Pix2pixHD~\cite{wang2018high}, CRN~\cite{chen2017photographic}, SIMS~\cite{qi2018semi} and GauGAN~\cite{park2019semantic}.
Results are shown in Fig.~\ref{fig:city1}, \ref{fig:city2}, \ref{fig:city3}, \ref{fig:ade20k_1} and \ref{fig:ade20k_2}.
We observe that the proposed LGGAN achieves visually better results than the competing methods. 

\subsection{Global and Local Generation}
\label{sec:3}
\begin{figure*}[!ht] 
	\centering
	\includegraphics[width=1\linewidth]{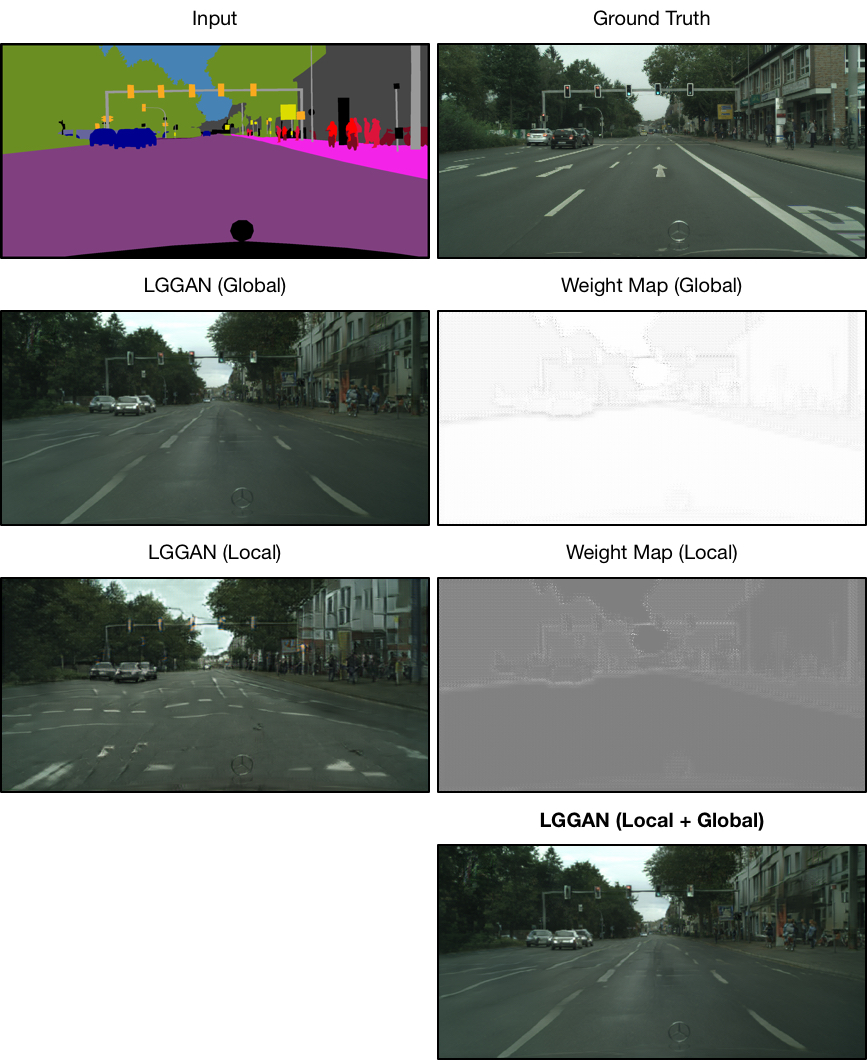}
	\caption{Results and weight maps generated by the proposed LGGAN with different settings on the Cityscapes dataset. These samples were randomly selected without cherry-picking for visualization purposes.
	}
	\label{fig:weight_map_city1}
\end{figure*}

\begin{figure*}[!ht] 
	\centering
	\includegraphics[width=1\linewidth]{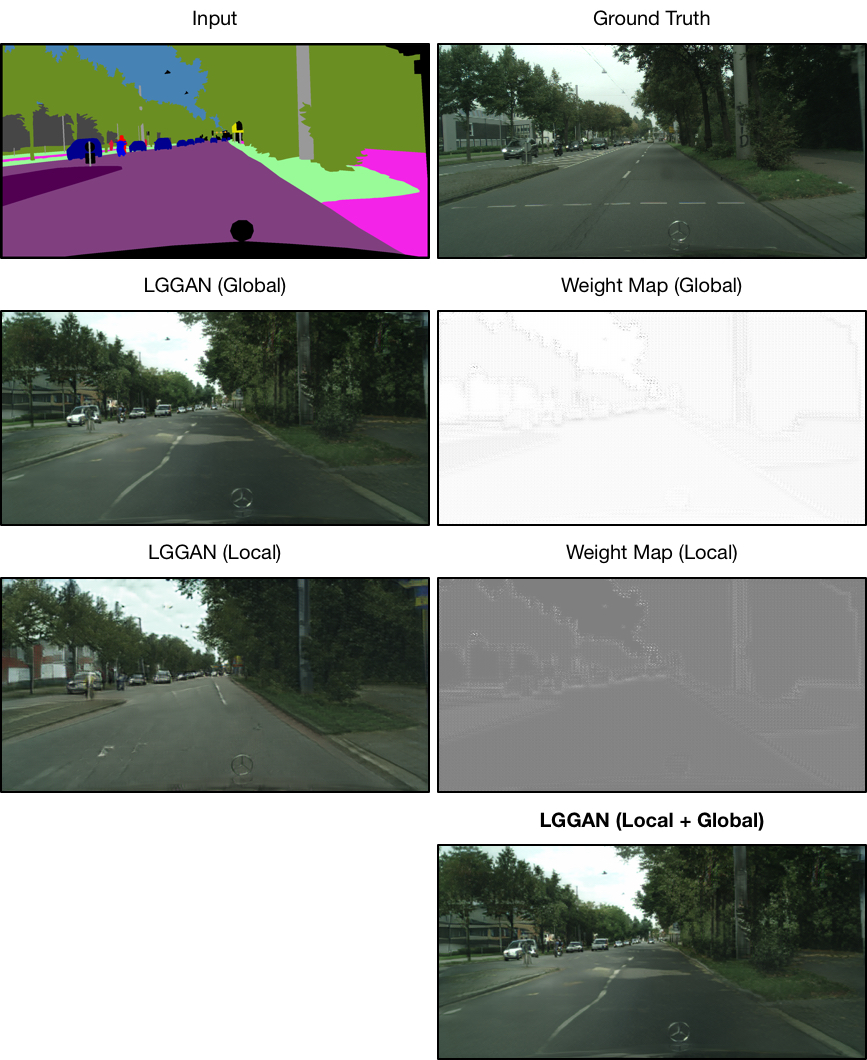}
	\caption{Results and weight maps generated by the proposed LGGAN with different settings on the Cityscapes dataset. These samples were randomly selected without cherry-picking for visualization purposes.
	}
	\label{fig:weight_map_city2}
\end{figure*}

\begin{figure*}[!ht] 
	\centering
	\includegraphics[width=1\linewidth]{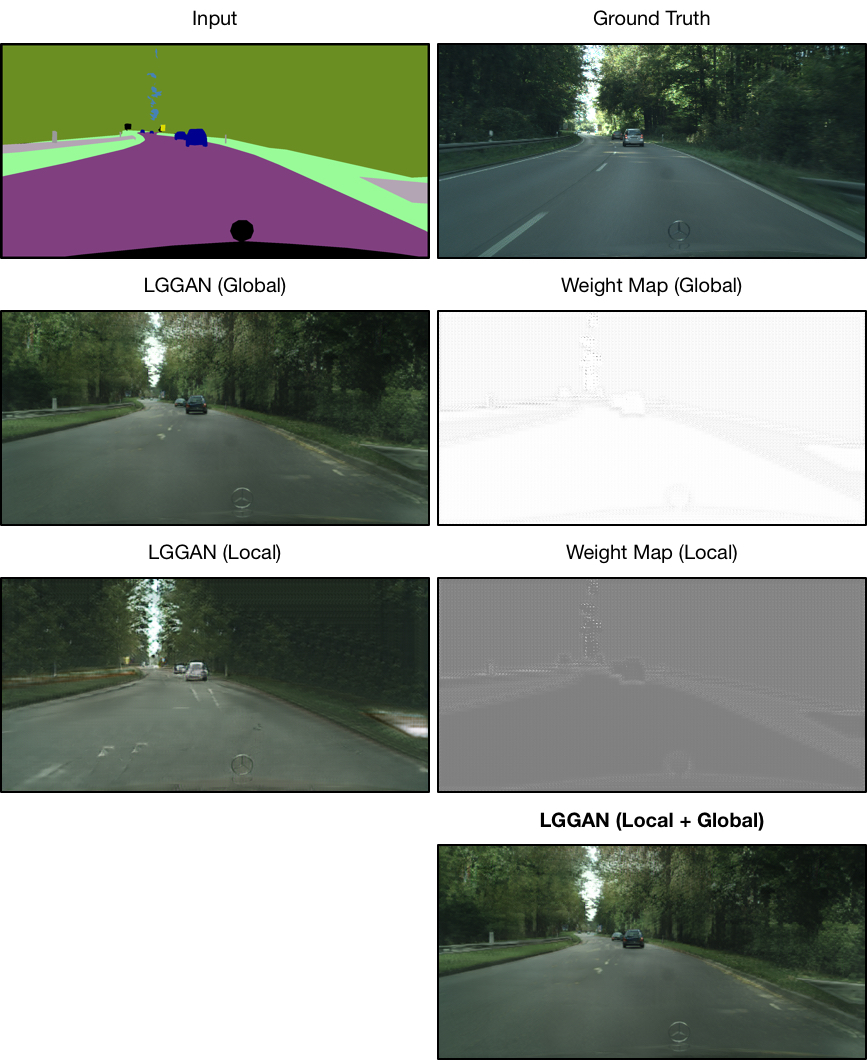}
	\caption{Results and weight maps generated by the proposed LGGAN with different settings on the Cityscapes dataset. These samples were randomly selected without cherry-picking for visualization purposes.
	}
	\label{fig:weight_map_city3}
\end{figure*}

\begin{figure*}[!ht] 
	\centering
	\includegraphics[width=1\linewidth]{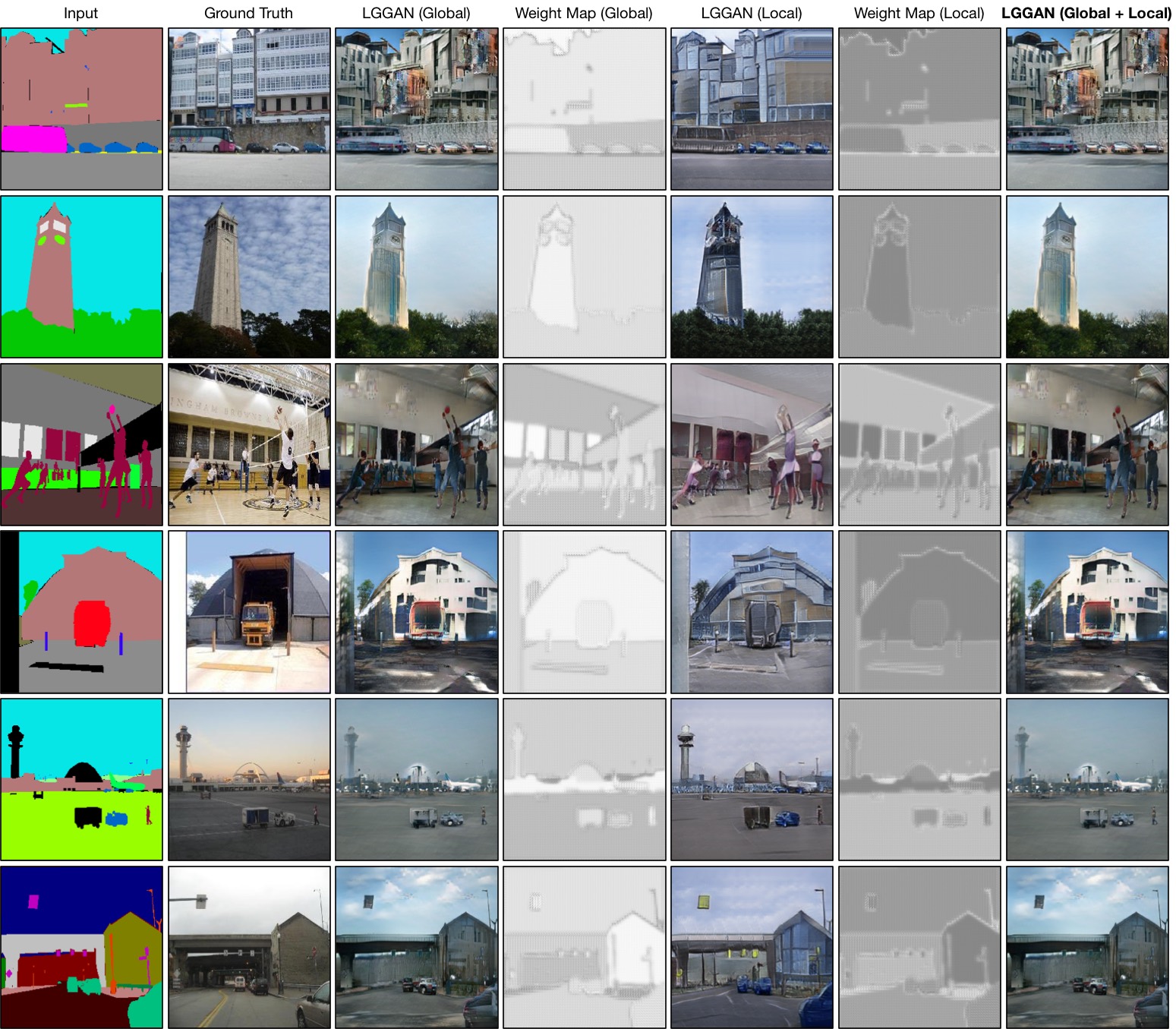}
	\caption{Results and weight maps generated by the proposed LGGAN with different settings on the ADE20K dataset. These samples were randomly selected without cherry-picking for visualization purposes.
	}
	\label{fig:weight_map1_20k}
\end{figure*}

\begin{figure*}[!ht] 
	\centering
	\includegraphics[width=1\linewidth]{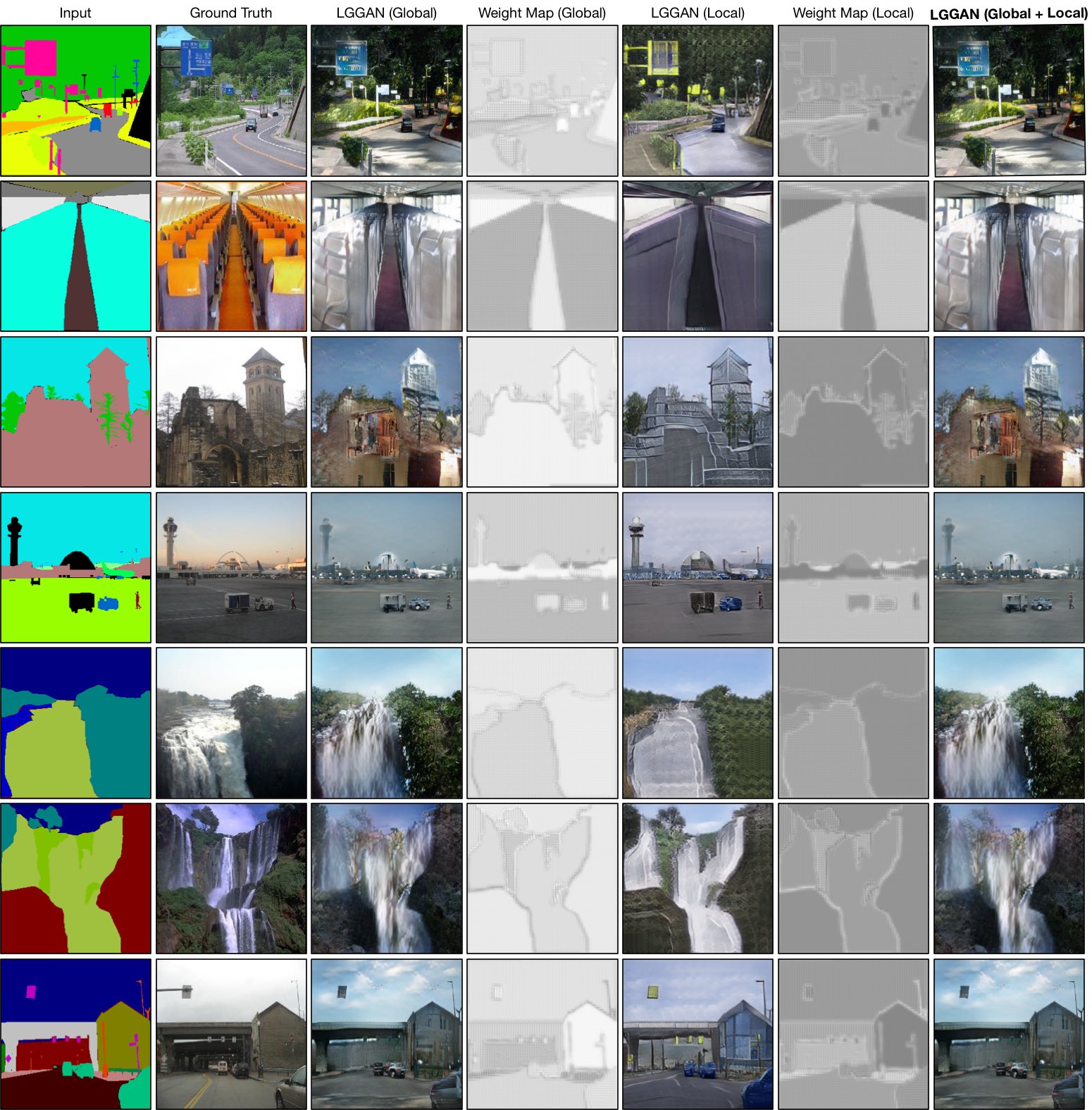}
	\caption{Results and weight maps generated by the proposed LGGAN with different settings on the ADE20K dataset. These samples were randomly selected without cherry-picking for visualization purposes.
	}
	\label{fig:weight_map2_20k}
\end{figure*}

We provide more comparison results of the proposed LGGAN with different settings on both datasets, i.e., `Local', `Global' and `Local + Global'. 
The qualitative generation results on the both datasets are shown in Fig.~\ref{fig:weight_map_city1}, \ref{fig:weight_map_city2}, \ref{fig:weight_map_city3}, \ref{fig:weight_map1_20k} and \ref{fig:weight_map2_20k}.
We observe that our full model (i.e., `Local + Global') generates visually much better results than both the local and the global models, further confirming our intuition to perform  joint adversarial learning on both the local and the global context.
Moreover, we also show the generated weight maps on these figures.
We observe that the generated global weight maps mainly focus on learning the global layout and structure, while the learned local weight maps focus on the local details, especially on the connection between different classes. 

\subsection{Visualization of Generated Semantic Maps}
\label{sec:6}
\begin{figure*}[!ht] 
	\centering
	\includegraphics[width=1\linewidth]{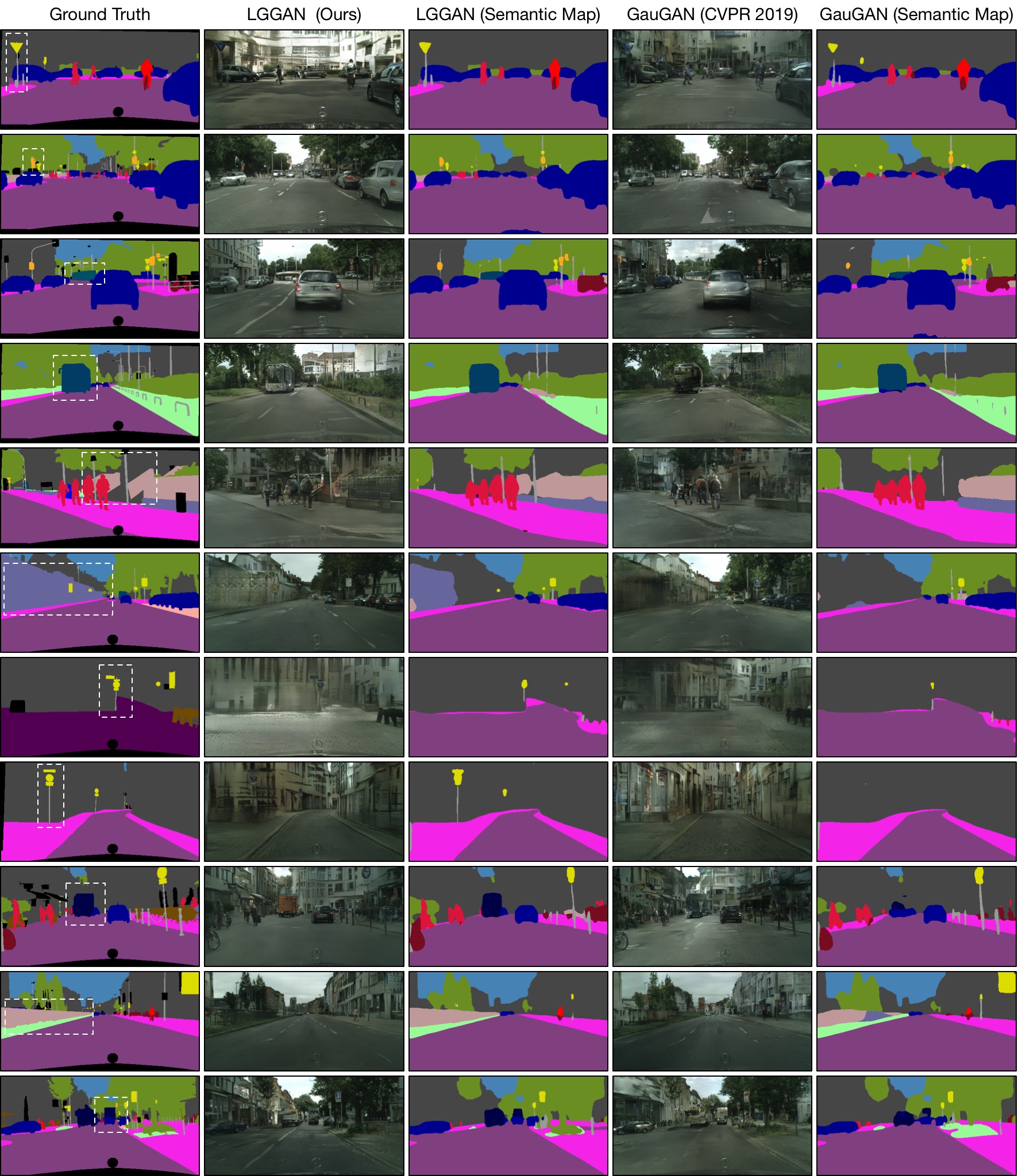}
	\caption{The generated semantic maps with comparison to those from GauGAN~\cite{park2019semantic} and ground truths on the Cityscapes dataset. These samples were randomly selected without cherry-picking for visualization purposes. Most improved regions are highlighted in the ground truths with white dash boxes. 
	}
	\label{fig:seg1}
\end{figure*}

\begin{figure*}[!ht] 
	\centering
	\includegraphics[width=1\linewidth]{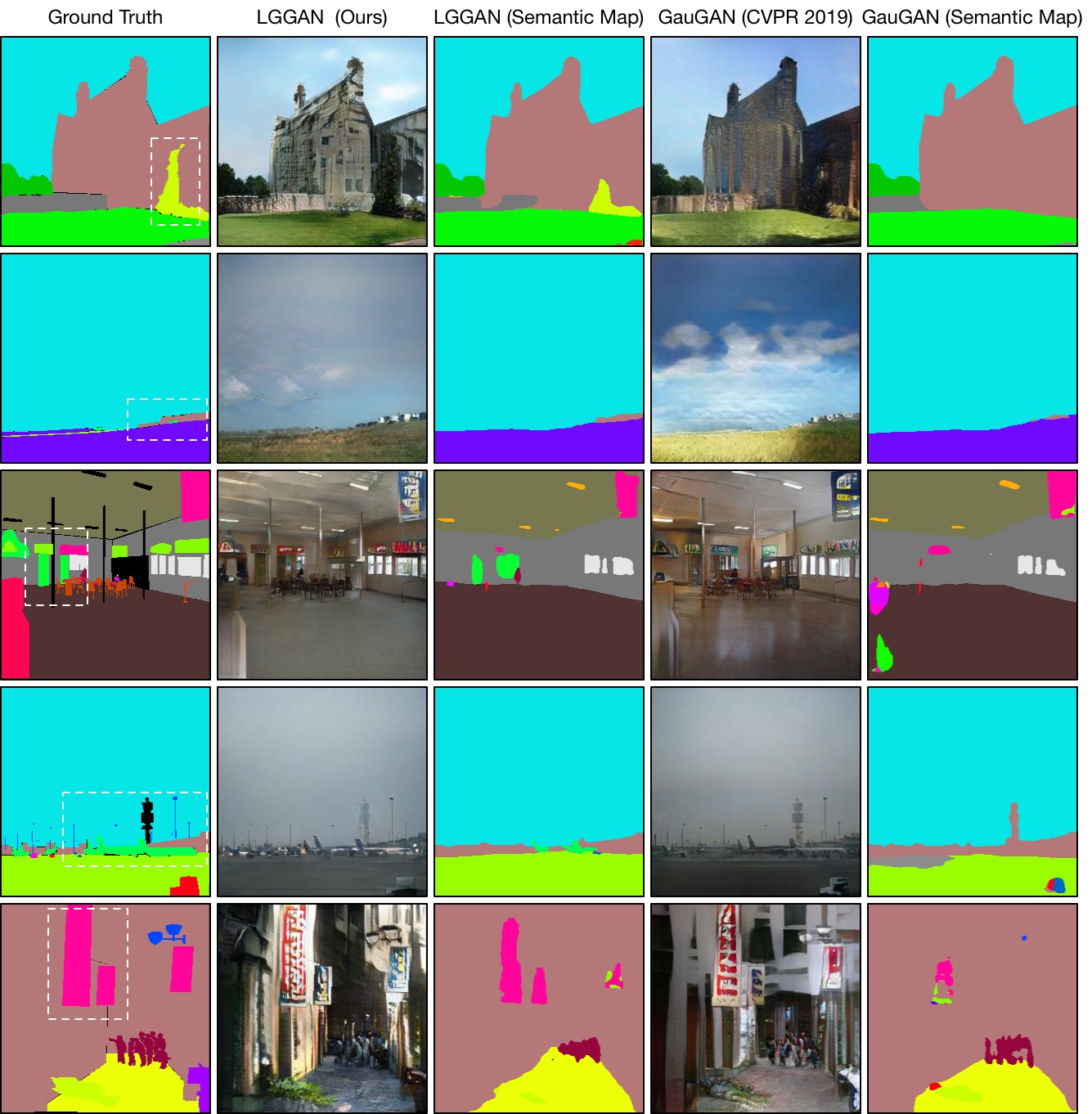}
	\caption{The generated semantic maps with comparison to those from GauGAN~\cite{park2019semantic} and ground truths on the ADE20K dataset. These samples were randomly selected without cherry-picking for visualization purposes. Most improved regions are highlighted in the ground truths with white dash boxes. 
	}
	\label{fig:seg1_20k}
\end{figure*}

\begin{figure*}[!ht] 
	\centering
	\includegraphics[width=1\linewidth]{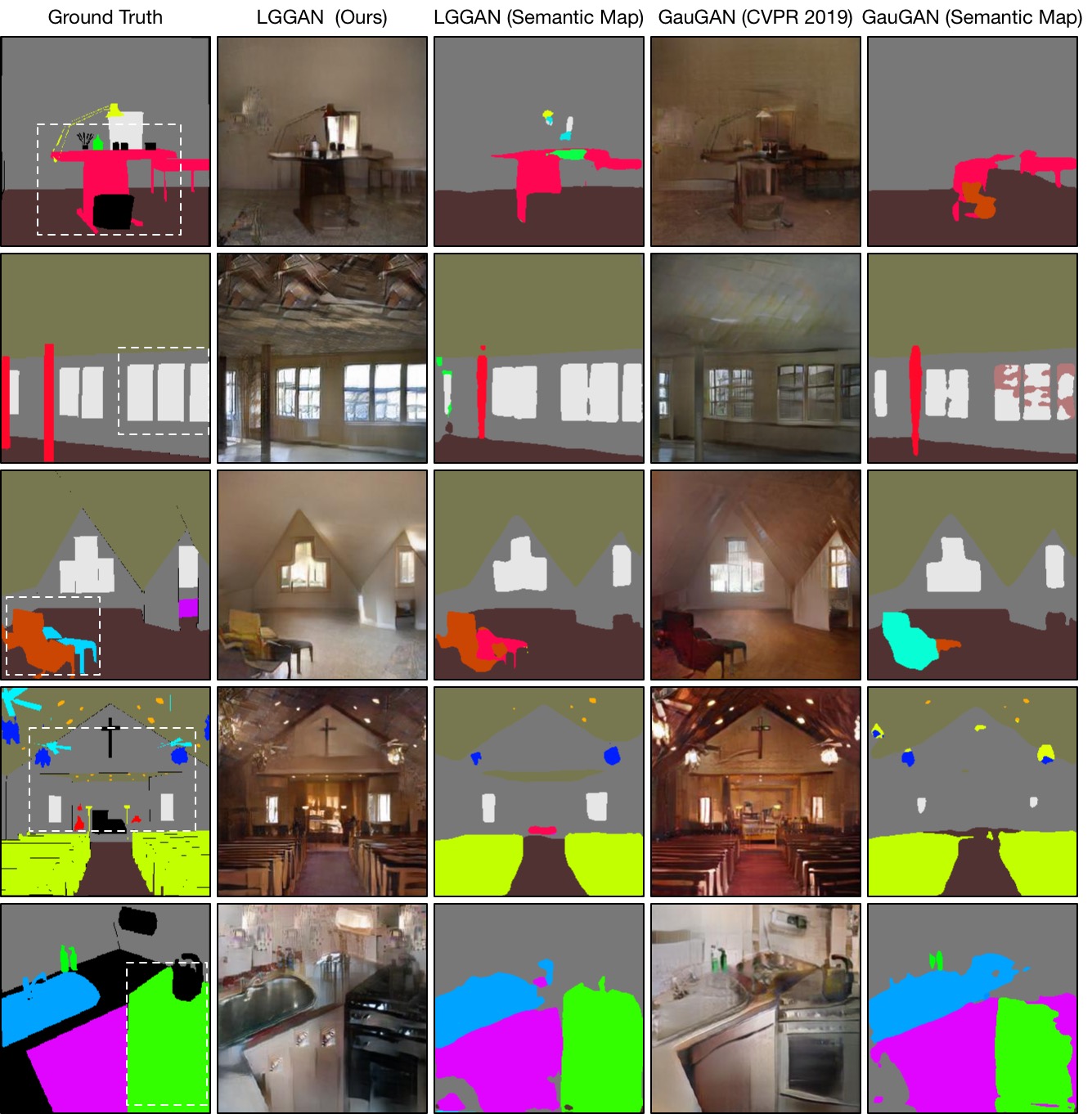}
	\caption{The generated semantic maps with comparison to those from GauGAN~\cite{park2019semantic} and ground truths on the ADE20K dataset. These samples were randomly selected without cherry-picking for visualization purposes. Most improved regions are highlighted in the ground truths with white dash boxes. 
	}
	\label{fig:seg2_20k}
\end{figure*}

We follow GauGAN~\cite{park2019semantic} and use
the state-of-the-art segmentation networks on the generated images to produce semantic maps:
DRN-D-105~\cite{yu2017dilated} for  Cityscapes and UperNet101~\cite{xiao2018unified} for ADE20K. 
The generated semantic maps of our LGGAN, GauGAN and the ground truth on both datasets are shown in Fig.~\ref{fig:seg1}, \ref{fig:seg1_20k} and~\ref{fig:seg2_20k}.
We observe that the proposed LGGAN
generates better semantic maps than GauGAN, especially
on local texture and small objects.

\subsection{Typical Failure Cases}
\label{sec:7}
\begin{figure*}[!ht] 
	\centering
	\includegraphics[width=1\linewidth]{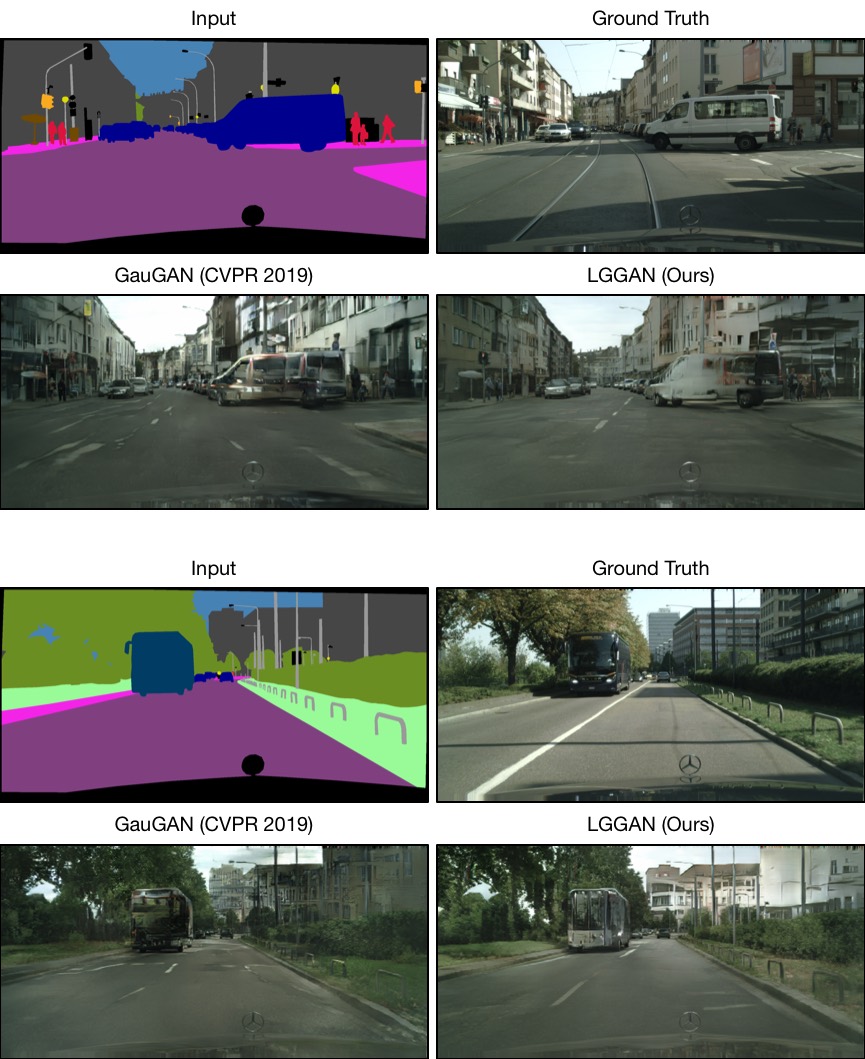}
	\caption{
	Failure cases with comparison to those from GauGAN~\cite{park2019semantic} and ground truths on the Cityscapes dataset.
	}
	\label{fig:case1}
\end{figure*}

\begin{figure*}[!ht] 
	\centering
	\includegraphics[width=1\linewidth]{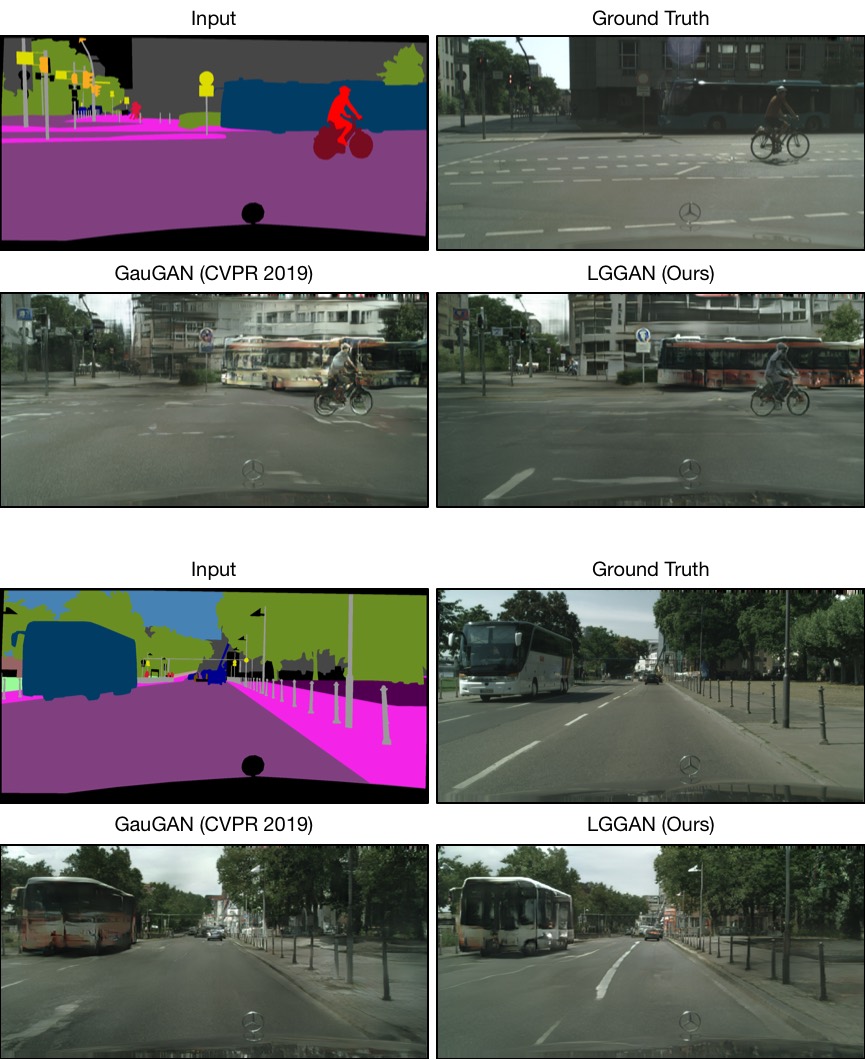}
	\caption{
		Failure cases with comparison to those from GauGAN~\cite{park2019semantic} and ground truths on the Cityscapes dataset.
	}
	\label{fig:case2}
\end{figure*}

\begin{figure*}[!ht] 
	\centering
	\includegraphics[width=1\linewidth]{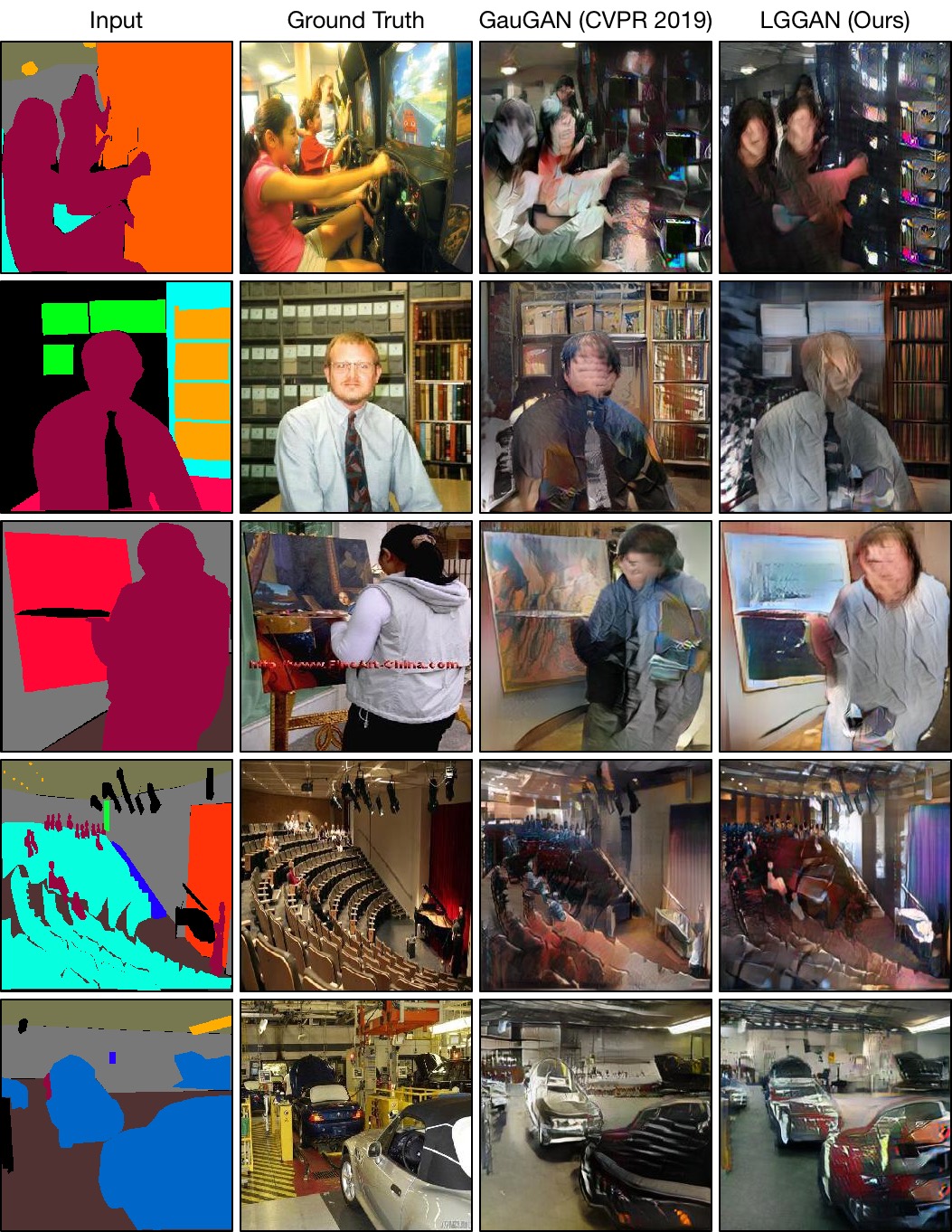}
	\caption{
		Failure cases with comparison to those from GauGAN~\cite{park2019semantic} and ground truths on the ADE20K dataset.
	}
	\label{fig:case3}
\end{figure*}

\begin{figure*}[!ht] 
	\centering
	\includegraphics[width=1\linewidth]{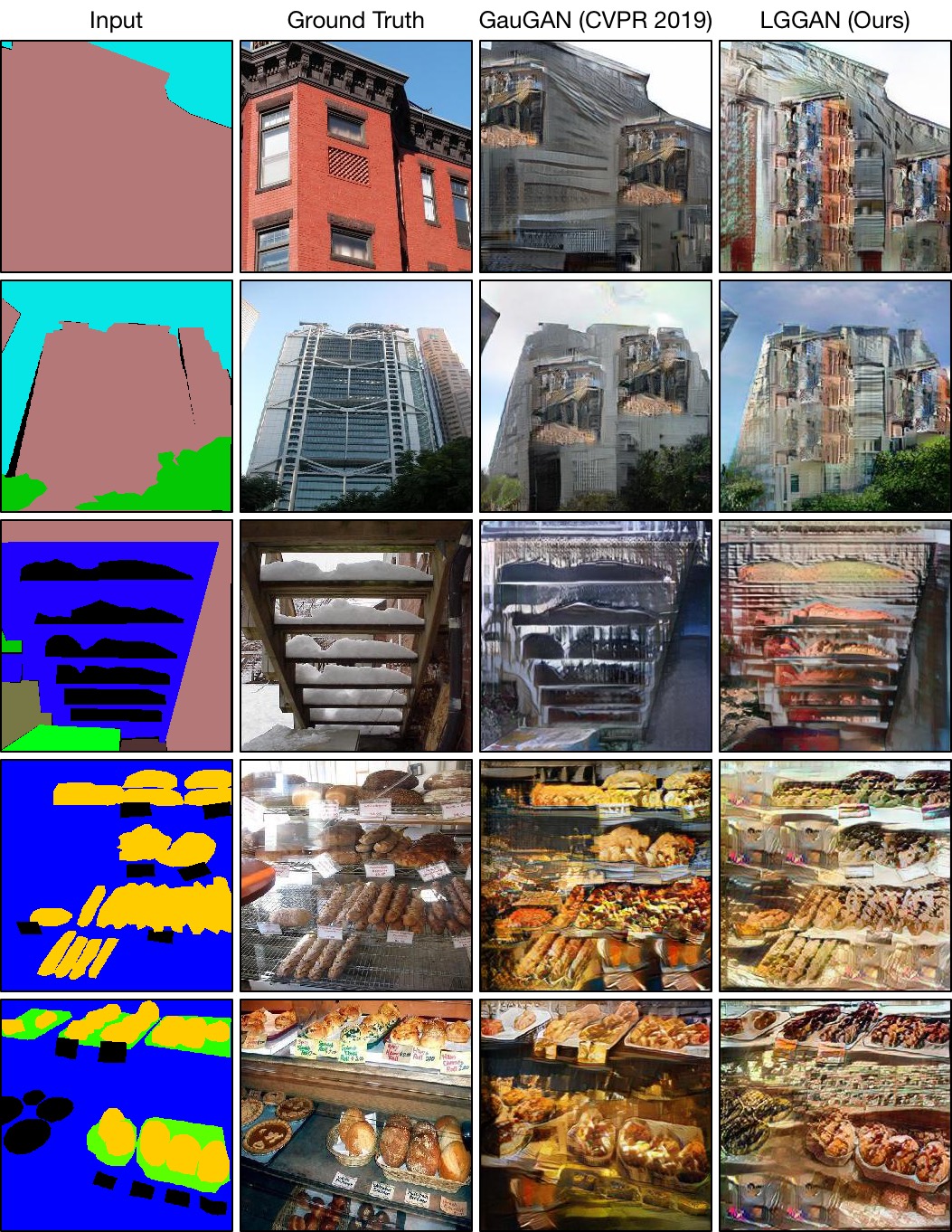}
	\caption{Failure cases with comparison to those from GauGAN~\cite{park2019semantic} and ground truths on the ADE20K dataset.
	}
	\label{fig:case4}
\end{figure*}

We also show some failure cases of our LGGAN compared with GauGAN~\cite{park2019semantic} on both datasets.
From Fig.~\ref{fig:case1} and \ref{fig:case2}, we can observe that the proposed LGGAN cannot generate photo-realistic object appearance on the Cityscapes dataset, especially large objects such as cars, trucks and buses.
From Fig.~\ref{fig:case3} and \ref{fig:case4}, we can see that the proposed LGGAN cannot generate several specific categories on the ADE20K dataset such as people, cars, houses and food.
The main reason for these failure cases could be the small number of samples in the training data.

%

